\documentclass[final,times,5p,twocolumn]{elsarticle}

\usepackage{mypreamble}

\begin{document}

    \begin{frontmatter}

\title{Hierarchical learning, forecasting coherent spatio-temporal individual and aggregated building loads}


\author[TUeaddress,DTUaddress]{Julien Leprince \corref{mycorrespondingauthor}}
\address[TUeaddress]{Technical University of Eindhoven, 5 Groene Loper, Eindhoven 5600 MB, the Netherlands}
\cortext[mycorrespondingauthor]{Corresponding author}
\ead{j.j.leprince@tue.nl}

\author[DTUaddress]{Henrik Madsen}
\address[DTUaddress]{Technical University of Denmark, Building 303B Matematiktorvet, Lyngby 2800, Denmark}
\ead{hmad@dtu.dk}

\author[DTUaddress]{Jan Kloppenborg M{\o}ller}
\ead{jkmo@dtu.dk}

\author[TUeaddress]{Wim Zeiler}
\ead{w.zeiler@tue.nl}

\begin{abstract}
Optimal decision-making compels us to anticipate the future at different horizons. However, in many domains connecting together predictions from multiple time horizons and abstractions levels across their organization becomes all the more important, else decision-makers would be planning using separate and possibly conflicting views of the future. 
This notably applies to smart grid operation. To optimally manage energy flows in such systems, accurate and \textit{coherent} predictions must be made across varying aggregation levels and horizons. Such hierarchical structures are said to be coherent when values at different scales are equal when brought to the same level, else would need to be reconciled.
With this work, we propose a novel multi-dimensional hierarchical forecasting method built upon structurally-informed machine-learning regressors and established hierarchical reconciliation taxonomy.
A generic formulation of multi-dimensional hierarchies, reconciling spatial and temporal hierarchies under a common frame is initially defined.
Next, a coherency-informed hierarchical learner is developed built upon a custom loss function leveraging optimal reconciliation methods.
Coherency of the produced hierarchical forecasts is then secured using similar reconciliation technics. The outcome is a unified and coherent forecast across all examined dimensions, granting decision-makers a common view of the future serving aligned decision-making.
The method is evaluated on two different case studies to predict building electrical loads across spatial, temporal, and spatio-temporal hierarchies, benchmarked against base and multi-task regressors.
Although the regressor natively profits from computationally efficient learning thanks to the unification of independent forecasts into a global multi-task model, results displayed disparate performances, demonstrating the value of hierarchical-coherent learning in only one setting. 
Yet, supported by a comprehensive result analysis, existing obstacles were clearly delineated, presenting distinct pathways for future work. Particularly, investigating reduced number of model weights and varying native multi-output regressors can tackle the two preeminent challenges that are the curse of dimensionality and coherency-learning complications from scaled trees respectively.
Overall, the paper expands and unites traditionally disjointed hierarchical forecasting methods providing a fertile route toward a novel generation of forecasting regressors.
\end{abstract}


\end{frontmatter}

    

    \section{Introduction}
A better anticipation of the future supports better decision-making. This is true across all sectors. Yet, more accurate forecasts alone often do not suffice. When dealing with different abstraction levels across a system or organization, it is commonly more important to obtain coherent predictions across all considered layers and horizons, not to result in unaligned decisions or possibly even conflicting ones \cite{nystrup2020temporal}.
This obstacle arises in multiple domains, including tourism \cite{kourentzes2019cross, athanasopoulos2009hierarchical}, retail \cite{kremer2016sum}, stock management \cite{spiliotis2021hierarchical} and smart grid management \cite{taieb2021hierarchical}, which showcases this matter quite adequately.

Traditionally, smart grid operators focused on forecasting the system's total demand. However, with the increasing adoption of smart meters at grid edges and substations, the focus is shifting. Grid management now benefits from high-frequency measurements available at multiple levels of aggregation allowing accurate forecast estimations across both spatial and temporal scales, i.e., from sub-meters to regional-level, with per seconds to monthly aggregated information \cite{taieb2021hierarchical, ahmad2020review, peng2021flexible}. Yet, the pluralities and independence of models and their consequent forecasts inevitably  produce inconsistencies across aggregation levels, i.e., lower-level predictions might not sum up to higher-level ones and vice-versa \cite{spiliotis2020cross}. 
The consequent challenge decision-makers are now faced with is to obtain coherent predictions across the different horizons and scales of the system. Hierarchical structures (or trees) are said to be coherent when their values at the disaggregate and aggregate scales are equal when brought to the same level \cite{kourentzes2019cross}. Should forecasts not be coherent, decision-making units would be planning using diverging views of the future. Optimal decision-making consequently requires forecasts to be coherent across all considered dimensional hierarchies.

\subsection{Hierarchical forecasting}
Enforcing coherency in hierarchical structures is a concept that dates back to 1942 \cite{stone1942precision} and was first defined in 1988 as \textit{reconciliation} \cite{weale1988reconciliation}. It leverages linear balancing equations from covariance compositions inherent to hierarchical structures to optimally re-adjust coherency mismatches. Hyndman et al. \cite{hyndman2011optimal} later reformulated the approach with a unifying statistical method, independent of prediction models, along with notations more appropriate to hierarchical forecasting.

Hierarchical forecasting can thus be defined as the process in which coherent predictions need to be made within a fixed hierarchical structure. 
Commonly, forecasts are first estimated separately considering each series of the hierarchy in a disjointed manner. These forecasts are designated as independent \textit{base} forecasts \cite{athanasopoulos2009hierarchical}. Generating base forecasts for each series implies that specialized models can be developed for each part of the hierarchy, incorporating node-specific available information \cite{kourentzes2019cross}.
Base forecasts are then linearly combined (reconciled) leveraging available information across the hierarchy to ensure coherency; a process employed by all hierarchical forecasting approaches as of to date \cite{nystrup2020temporal, athanasopoulos2009hierarchical, spiliotis2021hierarchical, taieb2021hierarchical, weale1988reconciliation, hyndman2011optimal, wickramasuriya2019optimal, 10.1007/978-3-319-18732-7_15, TIMMERMANN2006135, 8453006, ATHANASOPOULOS201760, spiliotis2019improving, bergsteinsson2021heat, nystrup2021dimensionality, yang2017reconciling}.

\subsubsection{Reconciliation approaches}
Predominant reconciliation techniques comprise traditional bottom-up and top-down approaches, trace minimization, optimal combinations, and recently developed machine-learning methods.

Bottom-up hierarchical forecasting consists in generating base forecasts at the very bottom level of the hierarchy and enforce coherency through their direct aggregation across the tree \cite{edwards1969should}.
The greatest advantage of this approach is that it can draw information from the most disaggregated levels of the tree, consequently avoiding any information loss from aggregation \cite{athanasopoulos2009hierarchical}. However, series located at tree leaves tend to possess low signal-to-noise ratios making them more difficult to predict. This is particularly true when dealing with smart-meter electrical demands which are notoriously volatile. Consumption peaks are indeed driven by often highly stochastic occupant behaviors that are close to intractable, consequently making bottom-up aggregation unlikely to provide accurate forecasts across the upper levels of the tree \cite{hyndman2011optimal}.

Top-down hierarchical forecasting on the other hand only generates forecasts for the top level of the hierarchy (tree-root) and proceeds to disaggregate and distribute it down the hierarchy from either historical \cite{grunfeld1960aggregation} or forecasted \cite{athanasopoulos2009hierarchical} proportions of the data. The approach commonly favors higher aggregation levels of the tree with more accurate predictions and is notably valuable for low-count data. However, aggregation is not without a large loss of information as temporal dynamics and other individual series characteristics cannot be exploited \cite{athanasopoulos2009hierarchical}. Additionally, as the success of this approach depends solely on one top-level model, it possesses a higher degree of risk from model misspecifications or inaccuracies \cite{kourentzes2019another}.
Given both bottom-up and top-down approaches inadequate to profit from the richness of information across a given hierarchy, optimal combination techniques emerged. Linearly reconciling base forecasts towards coherency, these approaches allowed interactions between different levels of the hierarchy, leveraging in particular correlations and covariances present in such structures \cite{hyndman2011optimal}. 

However, estimating the covariance structure of a hierarchy from base forecasts is challenging. Indeed, Wickramasuriya et al. \cite{wickramasuriya2019optimal} declared that the covariance matrix of the coherency errors is "impossible to estimate in practice due to identifiability conditions" such that even with high-frequency data available, assumptions on its form must be made \cite{doi:10.1080/01621459.2015.1058265}.
The ordinary least-square (OLS) estimator was particularly developed by Hyndman et al. \cite{hyndman2011optimal} and Athanasopoulos et al. \cite{athanasopoulos2009hierarchical} to avoid this problem. Their approach demonstrated improved results compared to other commonly adopted techniques.
A weighted least squares (WLS) approach, considering variances from the variance-covariance matrix diagonal but ignoring the off-diagonal covariance elements, was put forward by Hyndman et al. \cite{HYNDMAN201616}. Wickramasuriya et al. \cite{wickramasuriya2019optimal} later provided the theoretical justification for estimating variances from base forecast error variances. They proposed a generalized least-squares (GLS) estimator and found the incorporation of correlation information into the reconciliation process to benefit forecasting accuracy, with resulting reconciled forecasts guaranteed to be, in mean or in sample, at least, as good as their base forecasts, given a particular covariance structure.

Finally, in recent years, machine learning approaches have made their way into hierarchical forecasting. Relying on powerful statistical regressors and the availability of larger and richer data sets, machine learning emerges as an appealing and suitable tool for estimating the persistently challenging covariance matrix.
Spiliotis et al. \cite{spiliotis2021hierarchical} put forward such an approach employing a bottom-up method to reconcile predictions from Random Forest and XGBoost regressors. Taking as input the base forecasts of all the series of the hierarchy, the reconciled tree is then obtained from bottom-up aggregation. It allows non-linear combinations of the base forecasts, extending conventional linear approaches thanks to its machine-learning nature.
Sagheer et al. \cite{sagheer2021deep} proceeded to obtain coherent hierarchies from deep long-short term memory (DLSTM) recurrent neural networks by applying transfer learning across their hierarchies in a bottom-up fashion. They evaluated their approach on national-scale Brazilian electrical power production as well as Australian domestic tourism data.
In another work, Mancuso et al. \cite{mancuso2021machine} proposed a method to unify the two prevailing processes that are forecasting and reconciliation. By including hierarchical information in the forecasting process through a customized loss function, they allow the network to train towards reconciled forecasts using a top-down disaggregation process.

None of these approaches, however, include the general formulation of hierarchical forecasting within their learning framework. This  limits their reconciliation approaches to encompass solely traditional approaches, i.e., bottom-up or top-down, which, as has been mentioned, only exploit a fraction of the available information of hierarchical structures.

\subsubsection{Dimensional considerations}
While numerous works have first approached the reconciliation of hierarchical structures from a spatial (cross-sectional) dimensional frame perspective \cite{athanasopoulos2009hierarchical, spiliotis2021hierarchical, taieb2021hierarchical, weale1988reconciliation, hyndman2011optimal, wickramasuriya2019optimal, 10.1007/978-3-319-18732-7_15, TIMMERMANN2006135, 8453006}, temporal hierarchies have also been the center of recent attention within the field \cite{nystrup2020temporal, ATHANASOPOULOS201760, spiliotis2019improving, bergsteinsson2021heat, nystrup2021dimensionality, yang2017reconciling}.

Athanasopoulos et al. \cite{ATHANASOPOULOS201760} first introduced the notion of temporal hierarchies with forecasting reconciliation performed in the temporal dimension. Quite similarly to spatial reconciliation, base forecasts are independently produced across a defined set of temporal aggregation levels, e.g., weekly, daily, quarter-daily, hourly to per-minute or seconds granularities. This allows models to capture temporal-specific characteristics of the times series across the hierarchical-structure, e.g., trends or seasonality possessing particular time-frames.
Base temporal-forecasts are then reconciled across all forecasting horizons and temporal tree-structure, allowing aligned decisions across multiple planning horizons \cite{kourentzes2019cross}.
Nystrup et al. \cite{nystrup2020temporal} notably proposed temporal estimators accounting for autocorrelation structures to reconcile electric grid load forecasts. It was found that auto- and cross-covariances significantly improved forecast accuracy uniformly across all temporal aggregation levels.

It thus becomes clear that both spatial and temporal hierarchical forecasts produce substantial empirical accuracy improvements. By dealing with parameter estimation errors and model misspecifications, forecast combinations have demonstrated significant error variance reduction across numerous works \cite{KOURENTZES2016145, kourentzes2019another, winkler1992sensitivity}.
Exploiting both available hierarchical dimensions to further improve prediction accuracies consequently emerges as not only appealing but quite evident. 
Kourentzes and Athanasopoulos \cite{kourentzes2019cross} notably advanced a framework to produce spatial- and temporal-coherent forecasts (designated as cross-temporal), supporting all hierarchical levels with short- to long-term forecasts. Their work demonstrated empirical evidence that leveraging both dimensions in reconciliation offered improved accuracies compared to uni-dimensional reconciliation, i.e., spatial or temporal. A finding certainly due to the complete information exposure the approach provides.
Spiliotis et al. \cite{spiliotis2020cross} later proposed a cross-aggregation process to iteratively generate coherency across spatial hierarchies from multiple temporal aggregations applied to electricity consumption forecasting. Punia et al. \cite{punia2020cross} introduced a similar framework leveraging deep learning algorithms applied to supply chain base forecasts. Their approach, however, produced coherency solely from bottom-up approaches.

While the advantage of multi-dimensional hierarchical forecast has become evident, there exists, as of today, no generic formulation of these approaches. Indeed, while Spiliotis et al. \cite{spiliotis2020cross} stated that it is possible, in principle, to design a summing matrix S that accounts for both considered dimensions of reconciliation, a theoretical formulation of S and its subsequent reconciliation approaches was not put forward. Indeed, the design of a reconciliation estimator that fully captures scaling issues and cross-sectional interdependencies is not straightforward. Yet, this deprives multi-dimensional reconciliations of exploiting custom dimensional considerations.
The principal counterargument to undertaking such formulations is grounded on the fact that multi-dimensional hierarchies generate increasingly large tree structures that could soon become intractable to estimate.
Recently, however, the work of Nystrup et al. \cite{nystrup2021dimensionality} proposed a dimensionality reduction technique to counter this problem. Using eigendecomposition when reconciling forecasts, maximum information can be extracted from the error structure using available data. They find that uniformly improved predictions can be obtained across all aggregation levels, with the estimator achieving state-of-the-art accuracy all the while being applicable to hierarchies of all sizes.

\subsection{Motivation}
This comprehensive state-of-the-art overview underlines the following shortcomings;
\begin{enumerate}[label=(\roman*)]
    \item Base forecasts are typically produced separately, considering each series of the hierarchy in a disjointed manner. While this procedure allows the independence and hierarchically-tailored design of these models, it is inherently deprived from the benefits of data information (learning) transfer across models.
    \item Machine-learning reconciliation approaches have exhibited clear forecast improvements potential. Yet, developed approaches have, so far, not proceeded to put forward a unified method for machine-learning based hierarchical-forecasting. This limits considered reconciliations approaches to the more information-limited bottom-up and top-down approaches \cite{spiliotis2021hierarchical, punia2020cross}. Embedding advanced reconciliation techniques, e.g., optimal combinations, in the learning process of machine learning regressors is, as of today, still missing.
    \item Although advantages of leveraging multi-dimensional hierarchies in forecasting has become evident, a generic formulation of such hierarchical-combinations is still needed. Existing tools have demonstrated effective dimensionality reductions of large hierarchies \cite{nystrup2021dimensionality}, presenting promising solutions to the problem of dimension intractability.
\end{enumerate}

This study proposes a response to this appeal and puts forward a generic multi-dimensional formulation for hierarchical forecasting with machine-learning. 
We put together a unified and adaptable forecasting and reconciliation method founded on native multi-task machine-learning regressors while framing multi-dimensional hierarchical-forecasting approaches in a generic way. 
Contributions of this work can be summarized as five-fold;
\begin{enumerate}
    \item We develop a unified machine-learning-based hierarchical forecasting approach. This grants (\textit{i}) a unique forecasting model the benefit of a complete information overview across its hierarchy, while (\textit{ii}) including coherency constraints within its learning process as well as (\textit{iii}) being adaptable to either independent or combined forecasting and reconciliation processes. It establishes a unified method generating accurate and coherent forecasts at all levels of the hierarchy thanks to a custom hierarchical loss function leveraging coherency information from established field-taxonomy.
    \item To best exploit available information embedded within multi-dimensional data, we formulate a generic multi-dimensional extension of conventional hierarchical forecasting methods. In particular, we address the problem of diverging reconciliation considerations in a multi-dimensional setting with uni-dimensional couplings of the covariance estimator. This allows the unification of multi-hierarchical structures under a common frame, fueling both traditional and machine-learning approaches with ever-richer and transferable (learning) information.
    \item In the interest of addressing the dimensional tractability of our approach, we put forward dimensionality reduction prospects and illustrate them both theoretically and in practice with an applied demonstration.
    \item Our study considers two substantial smart-meter data sets including an established open source, i.e., the Building Data Genome project 2 (BDG2) \cite{miller2020building}. This allows the grounding of our approach thanks to a first-of-its-kind performance benchmark in the field of electric-meter hierarchical predictions, which we render fully replicable.
    \item To best serve knowledge dissemination and research reproducibility, we open-source all our developed code under the public GitHub repository \href{https://github.com/JulienLeprince/hierarchicallearning}{https://github.com/JulienLeprince/hierarchicallearning}.
\end{enumerate}

The greatest advantage of this approach is granting access to the regressor a complete information overview of the considered (multi-dimensional) hierarchy. This permits both a cross-dimensional, data-rich learning process as well as a hierarchically-informed training for hierarchical forecasting.
The outcome is a unified and coherent forecast across all examined dimensions, granting decision-makers a common view of the future serving aligned and better decisions.

The rest of this paper is organized as follows: Section \ref{sect_method} details traditional hierarchical forecasting prospects and extends them to multi-dimensional frames, while Section \ref{sec:hl} presents the hierarchical learning methods put forward with developed custom loss functions. Section \ref{sect_implementation} introduces the implementation specifics of our applied method from two different case studies, and Section \ref{sect_res} reports the performance results of the method. Highlighted findings are analyzed and detailed under the discussion Section \ref{sect_app}, followed by Section \ref{sect_con} which concludes the paper and reveals future work outlooks.

\section{Hierarchical forecasting}\label{sect_method}
In this section, we present the foundations of hierarchical forecasting as defined by Athanasopoulos et al. \cite{athanasopoulos2009hierarchical} and Wickramasuriya et al. \cite{wickramasuriya2019optimal} and extend them to multi-dimensional frames with a generic formulation. We discuss dimensionality tractability limitations and offer dimensionality reduction considerations to address them.

\subsection{Hierarchical structures}
\begin{figure}
    \centering
    \begin{adjustbox}{width=0.48\textwidth}
        \includegraphics{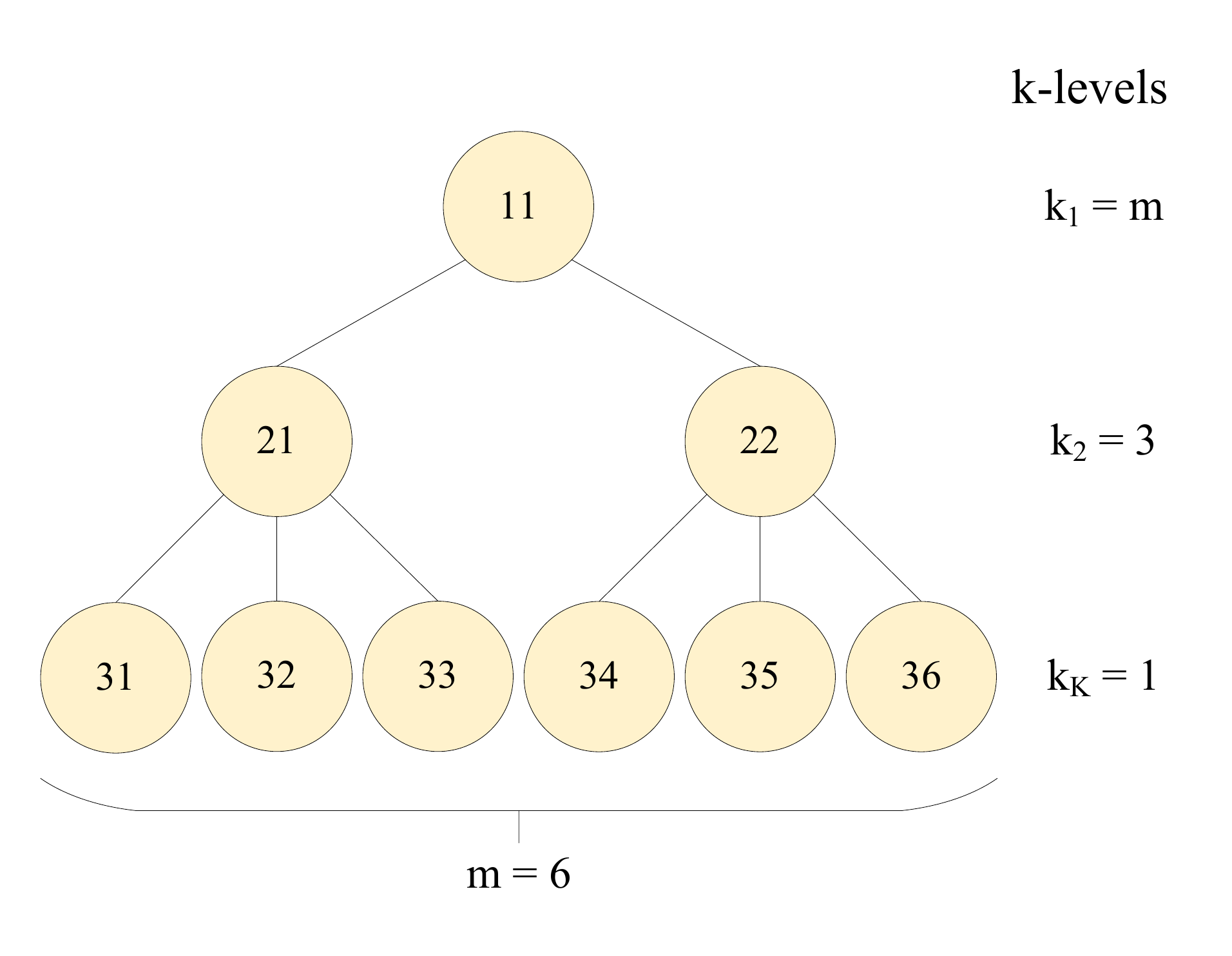}
    \end{adjustbox}
    \caption{A two-level hierarchical tree diagram.}
    \label{fig:tree}
\end{figure}

Let us refer to the simple hierarchy of Fig. \ref{fig:tree} to demonstrate the methodology. Every element (node) of the hierarchy (tree) can be labeled as $y_{kj}$, where the subscripts \textit{k} and \textit{j} stand for the aggregation-level and node observations respectively.
We define $k_1$ as the most aggregate level of the hierarchy (tree root), i.e., node $y_{11}$, and $k_K$ as the most disaggregate level (tree leaves), i.e., nodes $y_{Kj}$ where $j \in [1:m]$ and $K=3$.
In such a setting, two important components must be considered; the number of nodes in the bottom level of the hierarchical tree, which is denoted as \textit{m}, and the total number of nodes on the tree \textit{n}. Here $n=9$ and $m=6$.

Stacking all tree elements in a \textit{n}-dimensional vector $\boldsymbol{y}=(y_{11}, y_{21}, y_{22}, y_{31}, y_{32}, y_{33}, y_{34}, y_{35}, y_{36})^T$, and bottom-level observations in an \textit{m}-dimensional vector $\boldsymbol{b}=(y_{31}, y_{32}, y_{33}, y_{34}, y_{35}, y_{36})^T$, we can write
\begin{flalign}
    \boldsymbol{y} = S \boldsymbol{b} \text{,}
\end{flalign}
where $S$ is the summation matrix, here expressed as
\begin{flalign}
    S = 
    \begin{bmatrix}
    1 & 1 & 1 & 1 & 1 & 1 \\
    1 & 1 & 1 & 0 & 0 & 0\\
    0 & 0 & 0 & 1 & 1 & 1 \\
    \multicolumn{6}{c}{I_m} \\
    \end{bmatrix}
     \text{,}
\end{flalign}
which is of dimension $n \times m$, and $I_m$ is an identity matrix of size $m$.
$S$ maps the hierarchical structure of the tree, where from the tree leaves $\boldsymbol{b}$ the complete hierarchy $\boldsymbol{y}$ can be reproduced. 
Notice how $S$ captures the coherency requirements within the hierarchy, integrated here as the linear summations of the bottom-level observations.

\subsubsection{Uni-dimensional}
Hierarchical structures encompassed within hierarchical forecasting have, as of today, treated either one of the two following dimensional frames, namely, temporal $\mathcal{T}$ or spatial $\mathcal{S}$ (sectional).

We define \textit{spatial} dimensional perspectives as a unique inter-element dimension, which places itself in opposition to the previously-defined \textit{cross-sectional} dimensions \cite{kourentzes2019cross, spiliotis2020cross, ATHANASOPOULOS201760, di2021cross}, which aggregated elements from very different entities together, e.g., stock management, resulting in considerable heterogeneity within "one" (but in fact, multi-) dimension.
It is our proposal to re-frame these \textit{cross-sectional} considerations into separate dimensions to allow clear delineations of multi-dimensional frames, as we later detail in Sec. \ref{subsec:multidim_recon}.

Although structures of any shape or form can be designed in both dimensions, it is common for temporal hierarchies to adopt symmetrical structures, with k-level values being homogeneous across the trees' aggregation levels. 
Taking the exemplified symmetrical hierarchy of Fig. \ref{fig:tree}, one could consider removing nodes $y_{32}$ and $y_{33}$; resulting in a hierarchy where $m=4$, $n=7$ and node $y_{21}$ being consequently removed as a redundant element of $y_{31}$. This would result in a non-symmetrical tree which, in the temporal domain, implies non-equally spaced measurement points (or sampling rate) across the considered aggregation level and the ones above it.

Typically, for symmetrical trees, there are $k \in \{k_1 , . . . , k_K\}$ aggregation levels, where $k$ is a factor of $m$, with $k_1 = m$, $k_K = 1$, and $m / k$ is the number of observations at aggregation level k. The summation matrix of temporal hierarchies can therefore be expressed as \cite{nystrup2020temporal}
\begin{flalign}
    S_{\mathcal{T}} = 
    \begin{bmatrix}
    I_{m/k_1} \otimes  \boldsymbol{1}_{k_1}\\
    \vdots  \\
    I_{m/k_K} \otimes  \boldsymbol{1}_{k_K}\\
    \end{bmatrix}
     \text{,}
\end{flalign}
where $\otimes$ is the Kronecker product and $\boldsymbol{1}_k$ is a k-vector of ones.

To generically define the formulation of the summation matrix of any uni-dimensional hierarchy $\mathcal{H}$, however, one needs to consider the eventuality of non-homogeneous k-level values across aggregation levels as well as uneven tree-depths. To this end, we define
\begin{flalign}
    s_{ij} =
    \begin{cases}
        1,      & \text{if } y_{i} \hspace{3mm}\text{is ancestor of}\hspace{8mm} y_{Kj} \text{,}\\
        0,      & \text{if } y_{i} \hspace{3mm}\text{is not ancestor of}\hspace{3mm} y_{Kj} \text{,}
    \end{cases}
\end{flalign}
where $s_{ij}$ is a matrix element of the summation matrix $S_{\mathcal{H}}$ given a fixed hierarchical structure $\mathcal{H}$ and $y_{i}$ here refers to the $i$-th element of $\boldsymbol{y}$. The subscripts $i$ and $j$ go from 1 to $n-m$ and $m$ respectively. They refer to the considered tree node element $i$ and tree leaf element $j$.
This sets the matrix element of a given node $i$ to either 1 or 0 if it is an ancestor of the leaf element $j$, or, in other words, whether it is a result of the aggregation of the corresponding tree-leaf element $y_{Kj}$ or not respectively.
The summation matrix can then be expressed as
\begin{flalign}
    S_{\mathcal{H}} = 
    \begin{bmatrix}
    s_{11}      & \hdots &  s_{1j}      & \hdots    & s_{1m} \\
    \vdots      &        & \vdots       &           & \vdots \\
    s_{i1}      & \hdots & s_{ij}       & \hdots    & s_{im} \\
    \vdots      &        & \vdots       &           & \vdots \\
    s_{(n-m)1} & \hdots  & s_{(n-m)j}   & \hdots    & s_{(n-m)m} \\
    \multicolumn{5}{c}{I_m} \\
    \end{bmatrix}
     \text{.}
\end{flalign}

This enables the formulation of any hierarchical structure to a summation matrix, e.g., from event-based or equally spaced time-series measurements for temporal hierarchies $\mathcal{T}$, to non-symmetrical or homogeneous aggregation structures for spatial hierarchies $\mathcal{S}$.

\subsubsection{Multi-dimensional}\label{subsec:multidim_S}
\begin{figure*}
    \centering
    \begin{adjustbox}{width=0.98\textwidth}
        \includegraphics{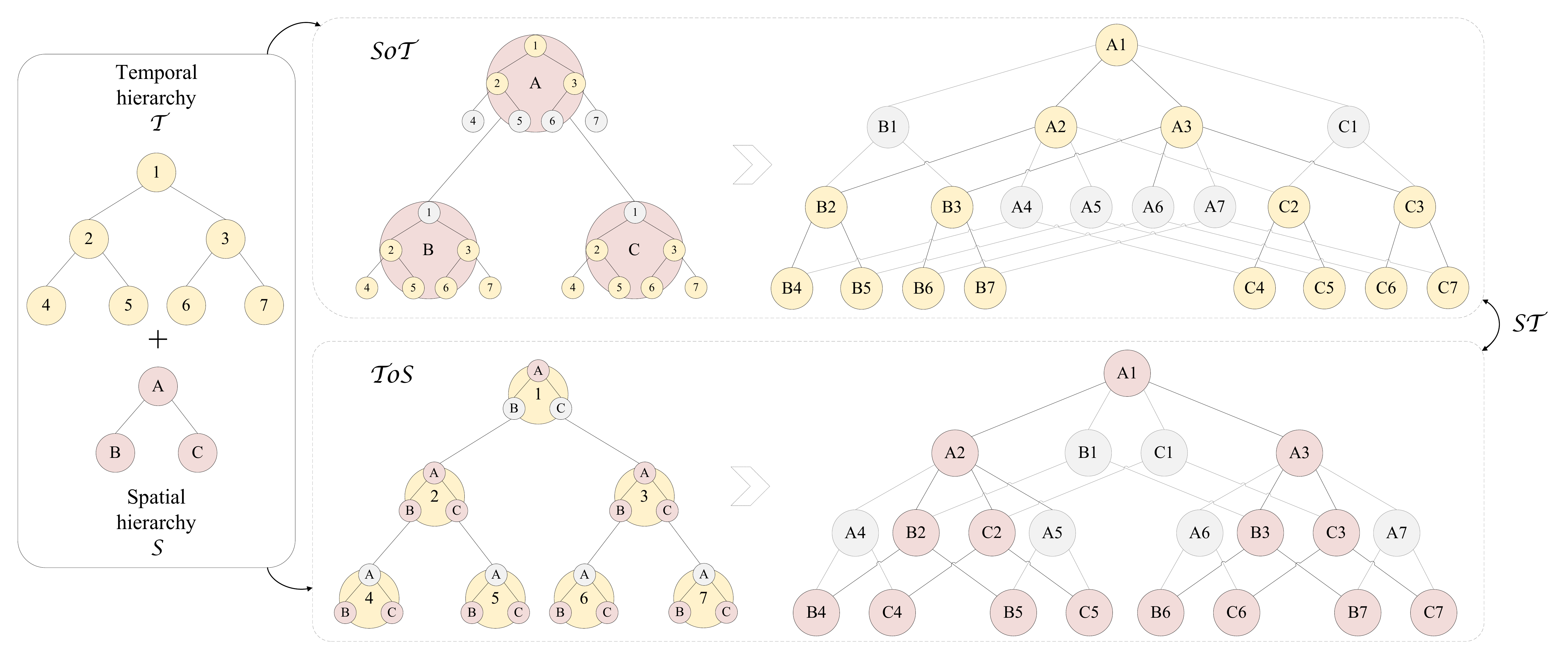}
    \end{adjustbox}
    \caption{Schematic of spatio-temporal $\mathcal{ST}$ hierarchical structure conception from either $\mathcal{S}o\mathcal{T}$ or $\mathcal{T}o\mathcal{S}$ structure composition, both producing an equivalent $\mathcal{ST}$ tree structures. Highlighted nodes (in grey) reveal opportunities for dimensionality reduction by dropping nodes of little dimensional interest, i.e., high temporal granularity in high spatial aggregation levels, and low temporal frequencies in high spatial granularities.}
    \label{fig:multitree}
\end{figure*}
\begin{figure*}
    \centering
    \begin{adjustbox}{width=0.98\textwidth}
        \includegraphics{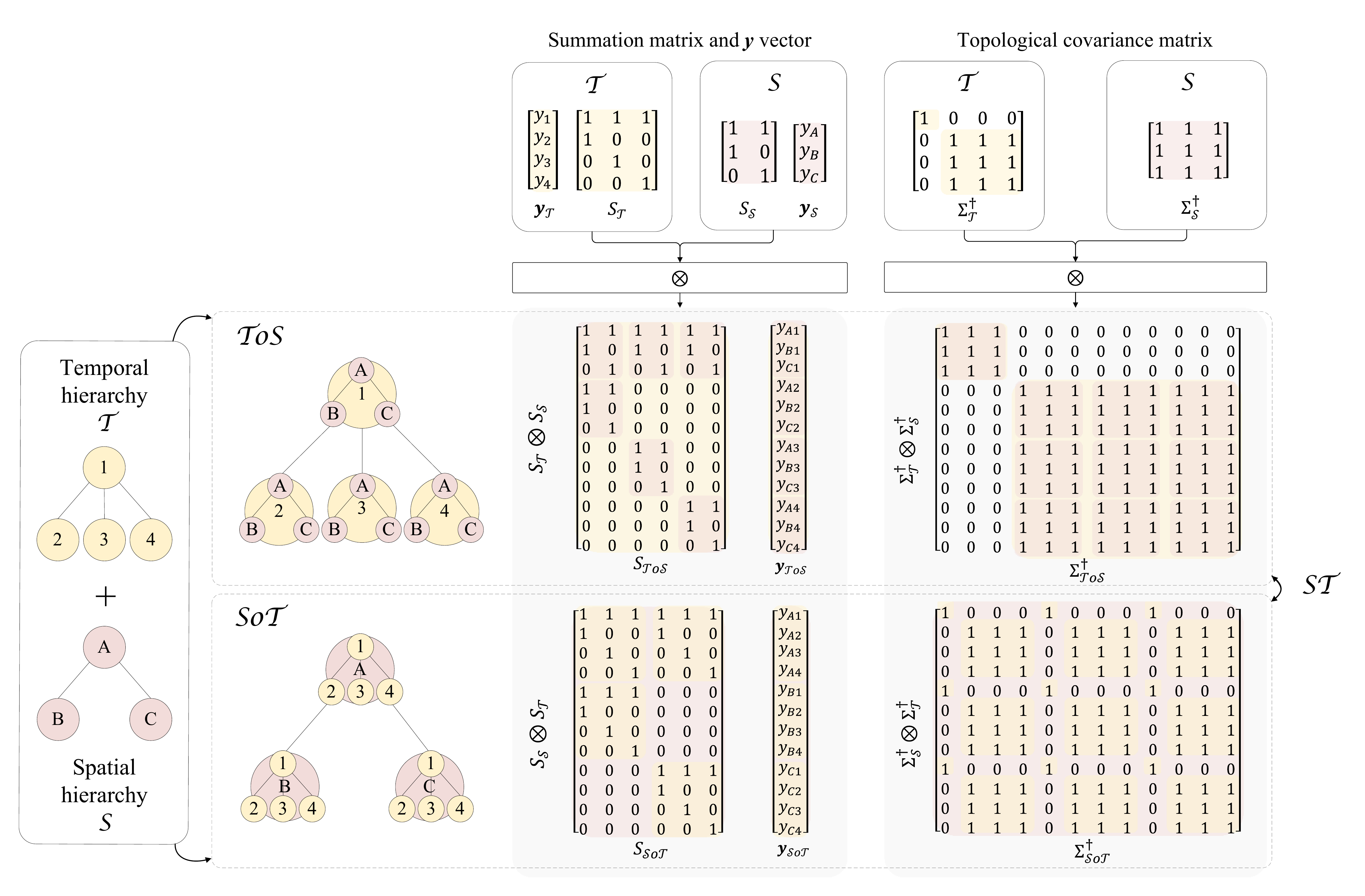}
    \end{adjustbox}
    \caption{Exemplified illustrations of hierarchical derivations of summation matrix, $\boldsymbol{y}$ vector and topological covariance matrix from spatio-temporal $\mathcal{S}o\mathcal{T}$ or $\mathcal{T}o\mathcal{S}$ function composition.}
    \label{fig:multitree_method}
\end{figure*}

Multi-dimensional hierarchies are the product of two uni-dimensional structures and can be obtained from function composition of separate hierarchical structures over another one. Fig. \ref{fig:multitree} illustrates the derivation of a spatio-temporal $\mathcal{ST}$ hierarchy from two disjointed spatial $\mathcal{S}$ and temporal $\mathcal{T}$ structure compositions, i.e., $\mathcal{S}o\mathcal{T}$ and $\mathcal{T}o\mathcal{S}$. The resulting tree structures demonstrate fundamental equivalences, with all tree nodes possessing identical bonds linking one element to the other, and consequently producing a unique hierarchical structure $\mathcal{ST}$.

The formulation of the multi-dimensional summation matrix in a generic way, can thus be expressed as a Kronecker product, where
\begin{flalign}
    S_{\mathcal{ST}} \equiv
    \begin{cases}
        S_{\mathcal{S}o\mathcal{T}} = & S_{\mathcal{S}} \otimes S_{\mathcal{T}} \text{,}\\
        S_{\mathcal{T}o\mathcal{S}} = & S_{\mathcal{T}} \otimes S_{\mathcal{S}} \text{,}\\
    \end{cases}
\end{flalign}
from which the resulting spatio-temporal summation matrix $S_{\mathcal{ST}}$ is of dimension $n_\mathcal{S}n_\mathcal{T} \times m_\mathcal{S}m_\mathcal{T}$, which, in the example of Fig. \ref{fig:multitree}, yields $3\cdot7 \times 2\cdot4 = 21 \times 8$.
The equivalence of $\mathcal{S}o\mathcal{T}$ and $\mathcal{T}o\mathcal{S}$ is attained via varying orderings of the $n_\mathcal{S}n_\mathcal{T}$-dimensional vector $\boldsymbol{y}_{\mathcal{ST}}$. 
These are derived from alternative transpose definitions of the observation matrix $Y_{{\mathcal{S}o\mathcal{T}}}$ such that
\begin{flalign}
    Y_{\mathcal{S}o\mathcal{T}} = Y_{\mathcal{T}o\mathcal{S}}^T =
    \begin{bmatrix}
    y_{11}      & \hdots & y_{1n_\mathcal{T}} \\
    \vdots      & \ddots & \vdots \\
    y_{n_\mathcal{S}1} & \hdots    & y_{n_\mathcal{S}n_\mathcal{T}} \\
    \end{bmatrix} \text{,}
\end{flalign}
where uni-dimensional vectors $\boldsymbol{y}_{\mathcal{S}}$ and $\boldsymbol{y}_{\mathcal{T}}$ are stacked together to form an observation matrix $Y_{\mathcal{S}o\mathcal{T}}$ of dimension $(n_\mathcal{S}, n_\mathcal{T})$. The $\boldsymbol{y}_{\mathcal{ST}}$ equivalent vectors can then obtained with
\begin{flalign}
    \boldsymbol{y}_{\mathcal{ST}} \equiv
    \begin{cases}
        \boldsymbol{y}_{\mathcal{S}o\mathcal{T}} = & \text{vec}(Y_{\mathcal{S}o\mathcal{T}}^T) \text{,}\\
        \boldsymbol{y}_{\mathcal{T}o\mathcal{S}} = & \text{vec}(Y_{\mathcal{T}o\mathcal{S}}^T) \text{.}\\
    \end{cases}
\end{flalign}
In the exemplified structures of Fig. \ref{fig:multitree}, we obtain $\boldsymbol{y}_{\mathcal{S}o\mathcal{T}} = (y_{A1}, y_{B1} , y_{C1}, ... , y_{A7}, y_{B7} , y_{C7})^T$ and $\boldsymbol{y}_{\mathcal{T}o\mathcal{S}} = (y_{A1}, ... , y_{A7}, y_{B1}, ... , y_{B7}, y_{C1}, ... , y_{C7})^T$.

With structural combinations of two disjointed dimensional hierarchies producing a unique bi-dimensional structure, it consequently follows that multi-dimensional combinations can be exploited in a similar manner. By chaining function compositions of considered \textit{singular} dimensions over summation matrices and $\boldsymbol{y}$ vectors, any combination of dimensional frames can be considered.

\subsubsection{Dimensionality reduction}
Multi-dimensional trees, however, introduce a key limitation: the dimensional explosion of hierarchical structures from function composition.
With the multiplication of dimensions from summation matrices, what was then considered a tractability shortcoming has now become an inevitable obstacle needing overcoming.

However, multi-dimensional hierarchies bring with them a consequential consideration: multi-dimensional aggregation levels. Indeed, such trees encompass more than former uni-dimensional high- or low-aggregation levels, they consist of deep structures where multi-dimensional aggregation combinations demand investigation. Spatio-temporal hierarchies, for example, display dissimilar insights from high-temporal-low-spatial aggregation levels, low-temporal-low-spatial or high-temporal-high-spatial ones.

It thus comes to light that, given a defined insight-driven application, subsets of certain multi-dimensional aggregation regions can be of limited use. High-frequency forecasts at very aggregate geographical levels might be of great value to grid operators contemplating frequency control in power systems, but not so much when forecasting tourism flows for instance \cite{kourentzes2019cross}.
Considering the end-goal application of optimal smart-grid control from electric load forecasting of grid edges (smart-building meter), low temporal frequencies and low spatial aggregations would be of little interest. Indeed, frequency control focuses on rather high-frequency samplings at medium-high spatial aggregation levels.
However, should the end-goal application be optimal cooperative control of smart-building neighborhoods, then low temporal frequencies and high spatial aggregations would become the dimensional frame of lesser concern.
Fig. \ref{fig:multitree} highlights these bi-dimensional nodes over the hierarchical structure conception, i.e., in grey, revealing the potential of dimensionality reduction within multi-dimensional hierarchies.

Therefore, while using spatio-temporal coherent forecasts offer benefits to decision-making, not all outputs from these hierarchies are effectively useful, opening the door to dimensionality reduction.

\subsection{Reconciliation methods}
Traditionally, forecast reconciliation starts by generating an initial forecast of the tree independently for each node, referred to as \textit{base} forecasts $\hat{\boldsymbol{y}}$. This set of hierarchical forecasts is stacked in the same manner as the $\boldsymbol{y}$ vector. Because of the independence of the base forecasts, in most cases, they do not exhibit coherency properties throughout their hierarchical structures. 
By introducing a matrix
\begin{flalign}\label{eq:G}
    G = \big[ 0_{m \hspace{.5mm}\times\hspace{.5mm} (n-m)} \hspace{.5mm}|\hspace{.5mm} I_m \big] \text{,}
\end{flalign}
of order $m \hspace{.5mm}\times\hspace{.5mm} n$ that extracts the $m$ bottom-level forecasts, the reconciliation constraint is formulated as
\begin{flalign}\label{eq:SG}
    \tilde{\boldsymbol{y}} = SG\tilde{\boldsymbol{y}} \text{.}
\end{flalign}
Reconciliation is necessary when base forecasts $\hat{\boldsymbol{y}}$ do not satisfy this constraint \cite{nystrup2020temporal}. In such situations, Eq. \eqref{eq:SG} becomes $\tilde{\boldsymbol{y}} = SG\hat{\boldsymbol{y}}$,
where $G$ maps the base forecasts into the reconciled tree-leaves and $S$ sums these up to a set of coherent forecasts $\tilde{\boldsymbol{y}}$. $SG$ can thus be thought of as a reconciliation matrix taking the incoherent base forecasts as input and reconciling them to $\tilde{\boldsymbol{y}}$.
A major drawback of traditional approaches is that $G$, as defined in Eq. \eqref{eq:G}, only considers information from a single level.

\subsubsection{Optimal reconciliation}

To include the exploitation of all aggregation levels in an optimal manner, Hyndman et al. \cite{hyndman2011optimal} and later, Van Erven and Cugliari \cite{10.1007/978-3-319-18732-7_15} and Athanasopoulos et al. \cite{ATHANASOPOULOS201760} formulated the reconciliation problem, as linear regression models. Exploiting either spatial or temporal hierarchical structures, reconciled forecasts are found employing the generalized least-squares estimate:
\begin{equation}\label{eq:optimalrecon2}
\begin{aligned}
    \text{minimize} &\hspace{5mm} \big(\tilde{\boldsymbol{y}} - \hat{\boldsymbol{y}}\big)^T \Sigma^{-1} \big(\tilde{\boldsymbol{y}} - \hat{\boldsymbol{y}}\big) \text{,}\\
    \text{subject to} &\hspace{5mm} \tilde{\boldsymbol{y}} = SG\tilde{\boldsymbol{y}} \text{,}
\end{aligned}
\end{equation}
where $\tilde{\boldsymbol{y}} \in \mathbb{R}^n$ is the decision variable of the optimization problem and $S \in \mathbb{R}^{n \times m}$ and $G  \in \mathbb{R}^{m \times n}$ are constant matrices defined by the structure of the hierarchy. The parameter $\Sigma \in \mathbb{R}^{n \times n}$ is the positive definite covariance matrix of the coherency errors $\varepsilon = \tilde{\boldsymbol{y}} - \hat{\boldsymbol{y}}$, which are assumed to be multivariate Gaussian and unbiased, i.e., with zero mean.

If $\Sigma$ were known, the solution to \eqref{eq:optimalrecon2} would be given by the generalized least-squares (GLS) estimator
\begin{flalign}\label{eq:sigma}
    \tilde{\boldsymbol{y}} = S \big(S^T \Sigma^{-1} S \big)^{-1} S^T\Sigma^{-1}\hat{\boldsymbol{y}} \text{,}
\end{flalign}
which has been employed in close to all notable hierarchical forecasting works over the last years \cite{nystrup2020temporal, kourentzes2019cross, athanasopoulos2009hierarchical, taieb2021hierarchical, spiliotis2020cross, hyndman2011optimal, wickramasuriya2019optimal, ATHANASOPOULOS201760}.
The precision matrix $\Sigma^{-1}$ is used to scale discrepancies from the base forecasts, hence, is often referred to as a weight matrix.

The recurrent challenge in estimating $\Sigma^{-1}$ stems from its dimension $n \times n$ which can potentially become very large.

\subsection{Multi-dimensional reconciliation}\label{subsec:multidim_recon}
Traditional uni-dimensional estimators can be coupled together topologically to form multi-dimensional ones in a similar manner to the summation matrix, with
\begin{flalign}
    \Sigma^{\dag}_{\mathcal{ST}} \equiv
    \begin{cases}
        \Sigma^{\dag}_{\mathcal{S}o\mathcal{T}} = & \Sigma^{\dag}_{\mathcal{S}} \otimes \Sigma^{\dag}_{\mathcal{T}} \text{,}\\
        \Sigma^{\dag}_{\mathcal{T}o\mathcal{S}} = & \Sigma^{\dag}_{\mathcal{T}} \otimes \Sigma^{\dag}_{\mathcal{S}} \text{,}\\
    \end{cases}
\end{flalign}
where $\Sigma^{\dag}$ refers to the topological covariance matrix of a given covariance matrix $\Sigma$. This allows uni-dimensional estimators $\Sigma_{\mathcal{S}}$ and $\Sigma_{\mathcal{T}}$ to incorporate dimension-specific topological considerations and produce a suitable multi-dimensional estimator $\Sigma_{\mathcal{ST}}$.

\begin{figure*}
    \centering
    \begin{adjustbox}{width=0.98\textwidth}
        \includegraphics{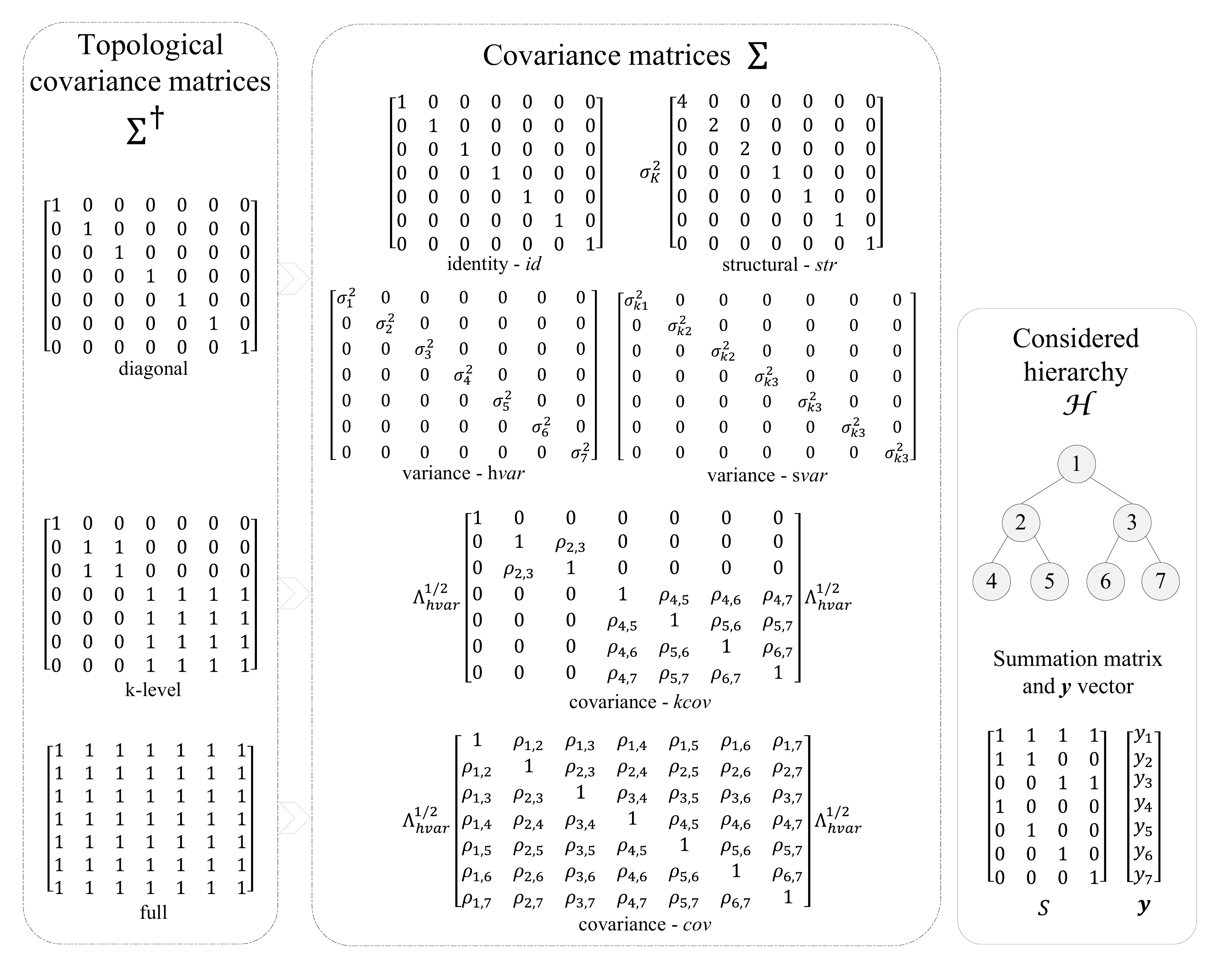}
    \end{adjustbox}
    \caption{Example illustration of the covariance matrices considered in this work along with their associated topological covariance matrices.}
    \label{fig:covariance_method}
\end{figure*}

The topological covariance matrix is characterized by elements of either 0 or 1 that indicate the mapping form assumption of the considered covariance matrix. Once the topological covariance matrix is identified, we simply populate it with the scaling parameters dictated by the reconciliation approach considered to obtain the covariance matrix.
Figure \ref{fig:multitree_method} exemplifies the identification of multidimensional topological covariance matrices from both $\mathcal{S}o\mathcal{T}$ and $\mathcal{T}o\mathcal{S}$ dimensional-derivations.

To address dimensional considerations in traditional estimators of the covariance matrix applied in reconciliation, we present four state-of-the-art estimators,  namely, identity, structural, variance, and covariance scaling with shrinkage, while detailing dimensional deliberations individually. 

\subsubsection{Identity}
A simplifying assumption proposed by Hyndman et al. \cite{hyndman2011optimal} puts the following identity approximation forward
\begin{flalign}\label{eq:sigma_id}
    \Sigma_{id} = I_{n} \text{.}
\end{flalign}
This simplistic approach has been shown to work well in practice \cite{athanasopoulos2009hierarchical} and allows to bypass the estimation of the covariance matrix. It ignores scale differences (captured by the variances) and interrelations (captured by the covariances) information of the observations within the hierarchical structure, which makes it independent of dimensional frame considerations.

Deep neural networks can be expected to build upon such simple relationships and approximate the more complex dependencies of the hierarchy thanks to its automated feature selection, as we later detail in Section \ref{sec:hl}.
We refer to this approach with the subscript \textit{id}.

\subsubsection{Structural scaling}
Structural scaling was proposed by Athanasopoulos et al. \cite{ATHANASOPOULOS201760} as a solution to cases where forecast errors are not available for some aggregation levels. It assumes the variance of each bottom-level base forecast error $\sigma^2_{K}$ is equal and that these are uncorrelated between nodes. Therefore, higher-level error variances are the sum of the error variances of tree leaves series connected to them.
By introducing a diagonal matrix $\Lambda_{str}$ with each element containing the number of forecast errors contributing to that aggregation level, they define
\begin{flalign}\label{eq:sigma_str}
    \Sigma_{str} &= \sigma^2_{K} \Lambda_{str} \text{,} \\
    \Lambda_{str} &= \text{diag}(S\boldsymbol{1}_m) \text{,}
\end{flalign}
where $\boldsymbol{1} \in \mathbb{R}^m$ is a column vector.
The hierarchy illustrated in Fig. \ref{fig:tree}, for instance, gives $\Lambda_{str} = \text{diag}(6, 3, 3, 1, 1, 1, 1, 1, 1)$.
The estimator is independent from the considered forecasting method, since no estimation of the variance of the forecast errors is needed, making it computationally efficient \cite{nystrup2020temporal}.

If considering a temporal dimensional frame, the estimator depends only on the seasonal period $m$ of the tree leaves. 
While with spatial perspectives, the estimator can suffer from heterogeneity within aggregation levels, e.g., residential and commercial buildings typically have heterogeneous electricity demand patterns and scale. Hence, assuming a common forecasting error variance across all leaves-series is not suitable \cite{kourentzes2019cross}.
We refer to this approach as \textit{str}.

\subsubsection{Variance scaling}
Another estimator proposed by Athanasopoulos et al. \cite{ATHANASOPOULOS201760}, referred to as variance scaling, scales the base forecasts using the variance of the residuals. It includes separate variance estimates for each aggregation level and assumes either homogeneous or non-homogeneous error variance within, but not across, a level.
Given the hierarchy presented in Fig. \ref{fig:tree}, this gives
\begin{flalign}\label{eq:sigma_var}
    \Sigma_{svar} &= \Lambda_{svar} = \text{diag}({\sigma}^2_{k_1}, {\sigma}^2_{k_2}, {\sigma}^2_{k_2}, {\sigma}^2_{k_3}, \ldots , {\sigma}^2_{k_3}) \text{,} \\
    \Sigma_{hvar} &= \Lambda_{hvar} = \text{diag}({\sigma}^2_{11}, {\sigma}^2_{21}, {\sigma}^2_{22}, {\sigma}^2_{31}, \ldots , {\sigma}^2_{37}) \text{,}
\end{flalign}
for homogeneous and heterogeneous variances respectively.
By definition, this scaling ignores correlations across and within aggregation levels and can be considered as an alternative weighted least-squares estimator.

Similarly to structural scaling, spatial and temporal dimensional scaling can differ due to their intrinsic heterogeneous and homogeneous error variances respectively; we refer to these estimators as \textit{hvar} and \textit{svar}. 
It follows, that $\Sigma_{svar}$ and $\Sigma_{hvar}$ are appropriate to temporal $\mathcal{T}$ and spatial $\mathcal{S}$ dimensional scalings respectively.

\subsubsection{Covariance scaling}
To exploit important information about a time series at different frequencies (temporal dimension) or inter-scale differences (spatial dimension), Nystrup et al. \cite{nystrup2020temporal} argue that potential information in the autocorrelation structure should be included.
They consequently proposed a \textit{covariance} scaling for temporal hierarchies estimating the full covariance matrix within each aggregation level, while ignoring correlations between them.

Following these footsteps, we explore both full and k-level, or so-called block, covariance estimates such that, for the hierarchy illustrated in Fig. \ref{fig:tree}, the estimator is either
\begin{flalign}
    \Sigma_{cov} = \Lambda^{1/2}_{hvar} R \Lambda^{1/2}_{hvar} \text{, or} \label{eq:sigma_cov} \\
        \Sigma_{kcov} = \Lambda^{1/2}_{hvar} R_{k} \Lambda^{1/2}_{hvar} \text{,} \label{eq:sigma_kcov}
\end{flalign}
where $R$ and $R_{k}$ refer to the full and k-level empirical cross-correlation matrix respectively,
\begin{flalign}
    R =
    \begin{bmatrix}
    1 & \hdots & \rho_{11,36} \\
    \vdots & \ddots & \vdots \\
    \rho_{11,36} & \hdots & 1 \\
    \end{bmatrix}
     \text{,}
\end{flalign}
\begin{flalign}
    R_{k} =
    \begin{bmatrix}
    1 & 0 & 0 & 0 & \hdots & 0 \\
    0 & 1 & \rho_{21,22} & 0 & \hdots & 0 \\
    0 & \rho_{22,21} & 1 & 0 & \hdots & 0 \\
    0 & 0 & 0 & 1 & \hdots & \rho_{31,36} \\
    \vdots & \vdots & \vdots & \vdots & \ddots & \vdots \\
    0 & 0 & 0 & \rho_{31,36} & \hdots & 1 \\
    \end{bmatrix}
     \text{.}
\end{flalign}


With increasing difficulties in estimating the full covariance matrix from high-dimensional hierarchies, even with high-frequency data available, special forms are commonly assumed. 
To alleviate this burden, Ledoit and Wolf \cite{ledoit2004well} proposed a Stein-type shrinkage estimator of the sample covariance matrix. Following these footsteps, Nystrup et al. \cite{nystrup2020temporal} considered a shrinkage estimator of the cross-correlation rather than the cross-covariance matrix to avoid problems with heteroscedasticity. Their estimator is based on decomposing the cross-covariance matrix into two diagonal (heterogeneous) variance matrices $\Lambda^{1/2}_{hvar}$ and a shrunk cross-correlation matrix $R_{srk}$.

The estimator is defined as
\begin{flalign}
    \Sigma_{srk} &= \Lambda_{hvar}^{1/2} R_{srk} \Lambda_{hvar}^{1/2} \text{,} \\
    R_{srk} & = (1 - \lambda)R + \lambda I_n \text{,} 
\end{flalign}
where $0  \leq \lambda \leq 1$ is a regularization parameter to control the degree of shrinkage towards the identity matrix.

When $\lambda = 1$, shrinkage scaling is equivalent to scaling by the diagonal variance matrix $\Lambda_{hvar}$. When $\lambda = 0$, it is equivalent to scaling by the sample covariance matrix.
A closed-form solution for the optimal value of $\lambda$ was derived by Ledoit and Wolf \cite{ledoit2004well} by minimizing the mean squared error. This shrinkage estimator is ideal for a small number of data points with a large number of parameters. With an assumed constant variance, the optimal shrinkage parameter is expressed by,
\begin{flalign}
    \lambda &= \frac{\Sigma_{i\neq j} \text{Var}(\sigma_{ij})}{\Sigma_{i\neq j} \sigma^2_{ij}} \text{,}
\end{flalign}
where $\sigma_{ij}$ is the $ij$th element of the covariance matrix from the base forecast errors. The variance of the estimated covariance, $\text{Var}(\sigma_{ij})$, is computed as depicted in Appendix A of Schäfer and Strimmer \cite{schafer2005shrinkage}.

Therefore, in contrast to the preceding variance and structural scaling estimators, this allows strong interrelations between time series in the hierarchy to be captured, while shrinkage alleviates the complexities of the estimation of $\Sigma_{srk}$ due to its size.

We refer to the shrunken estimators of Eqs. \eqref{eq:sigma_cov} and \eqref{eq:sigma_kcov} as \textit{cov} and \textit{kcov} respectively.

It should be noted that a variety of other well-performing estimators remain, including, but not limited to, Markov \cite{nystrup2020temporal} or spectral scaling \cite{nystrup2021dimensionality} supported by alternative inverse covariance shrinkage GLASSO method \cite{nystrup2020temporal}. In the intent of limiting the scope of this work to the evaluation of a novel hierarchical regressor, however, the afore-presented prevailing covariance approximation methods are favored. Figure \ref{fig:covariance_method} provides a visual illustration of the encompassed techniques along with their associated topological covariance matrices.

\subsection{Evaluation method}\label{subsec:evaluationmethod}
The accuracy evaluation of hierarchical forecasting performances requires the consideration of an important principle that common forecasting methods are exempt from, i.e., the structural scale differences inherent to hierarchical structures. Indeed, by its nature, hierarchical forecasting creates outputs of increasing orders of magnitudes, typically characterized by the aggregation levels of the tree, i.e., k-levels.
It consequently becomes crucial to take these hierarchically-impended scale differences into account when undertaking the accuracy performance evaluation of hierarchical forecasts, else these would consistently produce poorer performances for the top levels of the aggregation, where predicted values possess larger magnitudes.

This is commonly done by treating each aggregation level of the tree separately first, then evaluating the \textit{relative} per-level performance of the reconciliation phase over the base forecast, allowing the removal of scale differences between aggregation levels. However, a relative performance evaluation does not allow the comparison of approaches across case studies nor the distinctive performances of forecasting and reconciliation phases, which is why we propose to complement relative performance evaluations with measures based on structurally-scaled errors to provide an evaluation method more suited to the evaluation of hierarchical learning regressors.

\subsubsection{Relative measures}
The prevailing approach employed to evaluate hierarchical forecasting accuracy consists in scaling the accuracy performance of the reconciliation phase over a reference base forecast. This can be done by exploiting either Relative Mean Squared Error (RelMSE) \cite{kourentzes2019cross} or Relative Root Mean Squared Error (RRMSE) \cite{nystrup2020temporal, ATHANASOPOULOS201760, bergsteinsson2021heat}. Both depict the improvement of a given reconciliation approach compared to base forecast. 
We favor RelMSE over RRMSE to align with the commonly employed Mean Squared Error (MSE) loss function of machine learning models. The RelMSE can be expressed as
\begin{flalign}
    \text{RelMSE}_{k} &= \frac{\text{MSE}_{k}}{\text{MSE}^{base}_{k}} - 1 \text{,}
\end{flalign}
where the $\text{RelMSE}_{k}$ is computed for each aggregation level ${k}$ and $k \in \{1, 2, ..., K\}$. A negative entry describes a percentage improvement of the reconciled forecast over the base forecast. The $\text{MSE}_{k}$ is computed as the average error of all prediction steps of a given aggregation level ${k}$ from
\begin{flalign}
    \text{MSE}_{kj} &= \frac{1}{h}\sum^{h}_{t=1} e_{kj,t}^2 \text{,} \\
    \text{MSE}_{k} &= \frac{1}{N_k}\sum^{N_k}_{j=1} \text{MSE}_{kj} \text{,}
\end{flalign}
where $e_{kj,t} = y_{kj,t} - \hat{y}_{kj,t}$ is the forecast error at a starting reference time $t \in \mathbb{R}^h$ of an node $kj$ with $k$ being the aggregation level of the hierarchy possessing $N_k$ elements and $j$ the node observation. 
The starting reference time $t$ points to the very first time step considered in the hierarchy and is employed to anchor the nomenclature of temporal as well as spatio-temporal hierarchies, which usually encase time frames of [t, t+m], in similar notations as spatial ones.

\subsubsection{Measures based on scaled errors}
An alternative way of removing the inherent structural-scale differences present in hierarchical structures is producing structurally-scaled errors.
This can be achieved by dividing the error vector $\boldsymbol{e}_t = \boldsymbol{y}_t - \hat{\boldsymbol{y}}_t$ by the structural vector $\boldsymbol{\kappa}_{str}$, where each element contains the number of nodes contributing to the forecasted error of that aggregation level, such that
\begin{flalign}
    \boldsymbol{\kappa}_{str} &= S\boldsymbol{1}_m \text{,} \\
    \boldsymbol{e}^{str}_{t} &= \boldsymbol{e}_t \oslash \boldsymbol{\kappa}_{str} \text{,}\label{eq:streroor}
\end{flalign}
where $\oslash$ is a Hadamard division and $\boldsymbol{e}^{str}_{t}$ is the structurally scaled error vector at a time step $t$. The hierarchy illustrated in Fig. \ref{fig:tree}, for instance, gives $\boldsymbol{\kappa}_{str}$ = (6, 3, 3, 1, 1, 1, 1, 1, 1).

Structurally-scaled errors can then be employed in any given evaluation metric. We consequently define the Mean Structurally-Scaled Square Error (MS3E) as
\begin{flalign}
    \text{MS3E}_{kj} &= \frac{1}{h}\sum^{h}_{t=1} {e^{str}_{kj,t}}^2 \text{,}
\end{flalign}
which can be averaged either per aggregation level or over the entire hierarchy.

    \section{Hierarchical learning}\label{sec:hl}
\begin{figure*}
    \centering
    \begin{adjustbox}{width=0.99\textwidth}
        \includegraphics{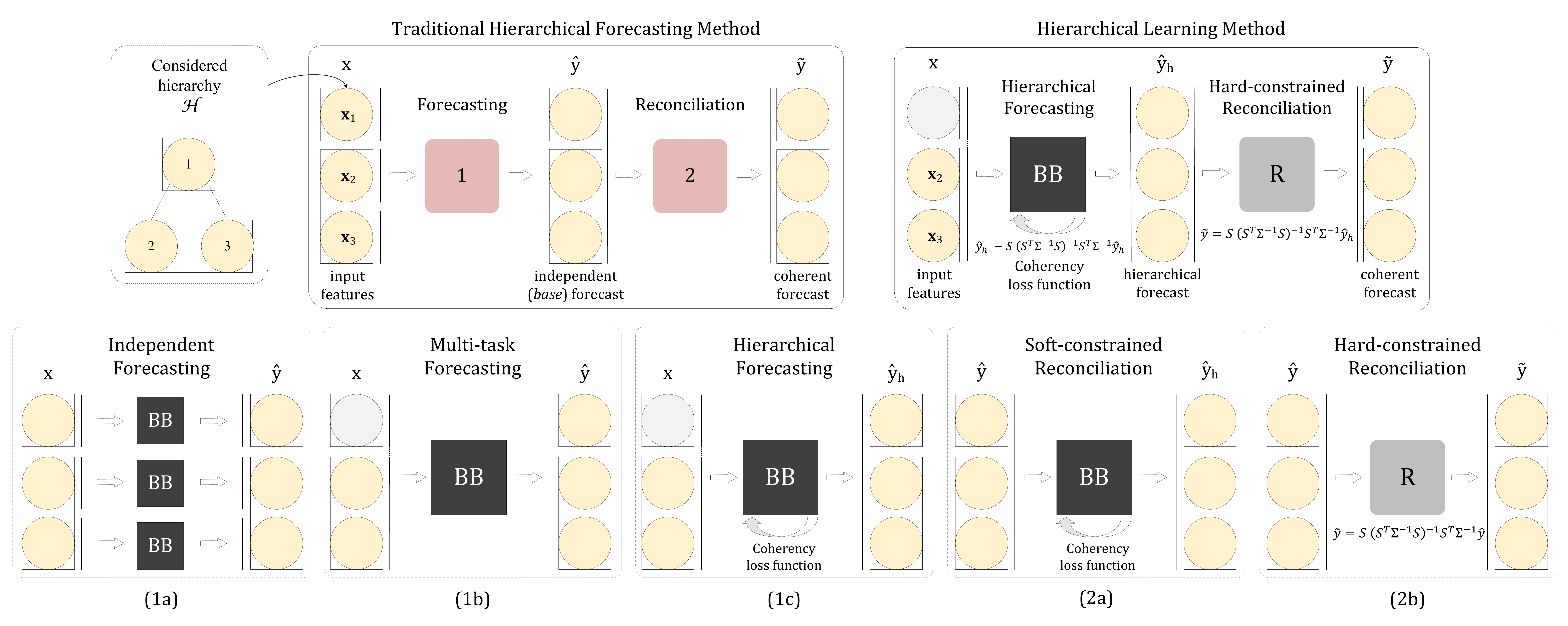}
    \end{adjustbox}
    \caption{Hierarchical forecasting methods with highlighted traditional approach steps (top left) and proposed hierarchical learning method (top right). Forecasting possibilities employing machine learning (here illustrated with the acronym BB, standing for Black Box) are illustrated (1a-c) along with two reconciliation approaches (2a-b). Forecasting methods encompass independent forecasting (1a), multi-task forecasting (1b), and our proposed hierarchical learning method (1c), working as a combined forecasting and reconciliation learner. Reconciliation approaches presented cover our machine learning method employed as a soft-constrained coherency enforcement over the base forecast (2a) and the traditional hard-constrained coherency enforcing method (2b).}
    \label{fig:hlearning}
\end{figure*}

While traditional hierarchical forecasting approaches have treated forecasting and reconciliation phases separately, we propose to unify these steps under a singular machine-learning method. 
To introduce our approach in a step-wise manner, let us first provide a comprehensive overview of the diverse ways machine learning may be employed within the frame of hierarchical forecasting, supported by the illustrative schemes provided in  Fig. \ref{fig:hlearning}.

We start by detailing the forecasting phase composing hierarchical forecasting with machine learning and continue with the description of the reconciliation step and its subsequent approaches.

\subsection{Hierarchical forecasting with machine learning}\label{sec:hierarchical_forecasting}
Machine learning regressors employed for hierarchical forecasting can here be employed in one of three ways.

\subsubsection{Independent forecasting}
First, with independent models each forecasting a unique node of the hierarchy, see (1a) of Fig. \ref{fig:hlearning}. The models leverage data information uniquely related to the considered node and do not exchange information with one another. They produce typical independent (\textit{base}) forecasts of the hierarchy.
Notable variations of this process involve either; 
(\textit{i}) exploiting transfer learning across the models to allow exchange of information throughout the hierarchy. The work of Sagheer et al. \cite{sagheer2021deep} precisely employed such a scheme using a top-down approach to determine coefficients of the lower-level models as proportions of the learnt top-level one. This process secures the coherency of the forecasted tree all the while providing privacy protection of data from one site to another, as transferred model coefficients retrieve data sharing dependence.
Or (\textit{ii}) by employing a unique single-output model for the forecasting of each node of the tree, namely a multivariate learner. This allows one model to gather more information as it learns from a much larger database than independent models. However, the disadvantage of this approach originates from the generalization intention of the learnt model applied to what could be, very different forecasted processes, e.g., heterogeneous buildings. This is why this approach works best when considering processes exhibiting similar characteristics, e.g., time series range, and typical patterns, which are generally obtained through a prior clustering phase \cite{leprince2020robust}. In addition, because the approach relies on the formulation of a unique model, any miss-specification could drastically impact the performance of the forecast, consequently making its design a key consideration for scientists.

The loss function of independent forecasting regressors are typically designed around a given error metric, e.g. mean squared error, describing the differences between forecasted and true values. Typically
\begin{flalign}\label{eq:Lb}
    \mathcal{L}^{b}\big(\mathcal{Y}, \widehat{\mathcal{Y}}|\Theta\big) = \frac{1}{h}\sum^{h}_{t=1} \big( y_t - \hat{y}_t \big)^2  \text{,}
\end{flalign}
where $\mathcal{L}^{b}$ denotes the mean square loss function between the predicted independent \textit{base} forecast set $\widehat{\mathcal{Y}}$ subject to a set of parameters $\Theta$ and a set of observed values $\mathcal{Y}$.

\subsubsection{Multi-task forecasting}
Second, by taking the concept of multivariate regressors even further, a multi-task regressor can be contemplated, see (1b) of Fig. \ref{fig:hlearning}. The regressor now produces a \textit{hierarchical}, dependent, forecast of the tree as a single vector output. The model notably accepts features from the bottom layer of the tree for spatial hierarchies, as aggregate levels would provide redundant information already present in the tree leaves. However, temporal hierarchies typically benefit from the inclusion of aggregate-level features, allowing them to exploit important information about the time series at different frequencies.
Requirements for coherency are, however, not included with such a scheme. 

The loss function of multi-task learners is similar to single-task ones other than considering vector rather than point errors, i.e.,
\begin{flalign}\label{eq:Lh} 
    \mathcal{L}^{h}\big(\mathcal{Y}, \widehat{\mathcal{Y}}|\Theta\big) = \frac{1}{h}\sum^{h}_{t=1} \Big( \boldsymbol{y}_t - \hat{\boldsymbol{y}}_t \Big)^2  \text{.}
\end{flalign}


\subsubsection{Hierarchical forecasting}
This takes us to our third and last approach, crystallizing the intention and concepts behind the contributions of our work, namely, hierarchical forecasting, see (1c) of Fig. \ref{fig:hlearning}. This technique builds on the aforementioned multi-task forecasting model while extending it with the inclusion of a coherency-informed learning process thanks to a custom loss function employing established coherency taxonomy from the literature.
The coherency loss function is formulated as the difference between the predicted values $\hat{\boldsymbol{y}}$ and its reconciled counterpart $\tilde{\boldsymbol{y}}$, following the reconciliation constraint of Eq. \eqref{eq:sigma}.
The coherency loss function $\mathcal{L}^{c}$ can consequently be expressed as
\begin{flalign}\label{eq:Lc}
    \mathcal{L}^{c}\big(\mathcal{Y}, \widehat{\mathcal{Y}}|\Theta\big) = \frac{1}{h}\sum^{h}_{t=1} \Big( \hat{\boldsymbol{y}}_t - S \big(S^T \Sigma^{-1} S \big)^{-1} S^T\Sigma^{-1}\hat{\boldsymbol{y}}_t \Big)^2 \text{.}
\end{flalign}
To combine both accuracy and coherency in the learning process of the regressor, the coherency loss is added to the hierarchical loss function defined in Eq. \eqref{eq:Lh} forming the hierarchical-coherent loss function $\mathcal{L}^{hc}$,
\begin{flalign}\label{eq:Lhc}
    \mathcal{L}^{hc}\big(\mathcal{Y}, \widehat{\mathcal{Y}}|\Theta\big) = \alpha \mathcal{L}^{h}_t + (1-\alpha) \mathcal{L}^{c}_t  \text{,}
\end{flalign}
where $\alpha \in [0,1]$ weights the hierarchical loss against the coherency loss. This avoids the over-adjustment of weights during the training of the regressor due to the addition of the coherency loss to the loss function. We typically set $\alpha$ to 0.75 for hierarchical forecasting to favor accuracy learning of produced predictions over coherency, yet this parameter should commonly be tuned by hyper-parameter optimization in the validation process of the model development, see Sec. \ref{subsubsec:modeldesign} for implementation details.

The method regroups numerous key advantages of machine learning-based forecasts. With large and rich multi-dimensional data to learn from the regressor effectively makes use of all the information provided by the most detailed layer of the hierarchy, i.e., the tree leaves, all the while incorporating hierarchical structure information as a soft-constrained learning mechanism.
Loss function augmentation via regularization and penalty methods has grown to become the most popular way of introducing constraints in deep learning \cite{pathak2015constrained, jia2017constrained, liu2018constrained}. Although the approach comes at the price of sacrificing hard constraints, it has been shown that soft-constrained penalty methods perform well in practice and often exceed hard constraint methods \cite{drgovna2021physics, marquez2017imposing}.
In addition, machine learning approaches are powerful at capturing non-linear relationships in the targeted predicted values. In particular, deep-learning methods are known for effective and automatic feature extraction from the data, thus reducing the need for guesswork and heuristics, which could provide a much-needed solution to the problem of non-identifiability of the covariance matrix.

Its disadvantages are similar to those of hierarchical forecasting approaches. By relying on a unique model, architecture considerations become paramount for the accurate performance of the regressor and consequently require careful, tailored tuning, e.g. with hyper-parameter grid-search.

\subsection{Reconciliation with machine learning}
While our proposed hierarchical learning approach (1c) blurs the limit between the traditionally delineated forecasting and reconciliation steps, it can also be employed as a classic reconciliation step, see (2a) of Fig. \ref{fig:hlearning}. 
Proposed as a soft-constrained coherency regressor, the machine learning model now takes the entire \textit{base} forecast $\hat{\boldsymbol{y}}$ as input and outputs a coherency-informed forecast $\hat{\boldsymbol{y}}_h$. The weighting coefficient $\alpha$ presented in Eq. \eqref{eq:Lc} can here be set to 0.25 to favor coherent outputs for example.
The evaluation of such a scheme, lays, however, outside the scope of this work, as our contribution targets hierarchical forecasting performance evaluation on varying dimensions. This setup rather showcases the flexibility of our approach as applicable to both the forecasting and reconciliation phases of traditional hierarchical forecasting methods.

For hard-constrained reconciliation, optimal reconciliation is considered, see (2b) of Fig. \ref{fig:hlearning}. It imposes coherency to its input forecast and can be employed a posteriori to the hierarchical learning step (1c) for eventual non-coherent outputs. In addition, as an established reconciliation method, it provides a good benchmark to evaluate the performance of our proposed method to both forecasting (1c) and reconciliation (2a).

    \section{Implementation}\label{sect_implementation}
This section details the implementation-related details of our study, namely, considered case studies, hierarchical structures, and predictive-learning setup.

\subsection{Case studies}\label{sect_casestudies}
Our study considers two large datasets of building smart-meter measurements to demonstrate the usability and performance of our method to real-life scenarios. 

Case Study 1 considers a total of 225 homes located in the Netherlands, a European region under the K\"oppen climate classification index \cite{CHEN201369} \textit{Cfb} which describes mild temperate, fully humid and warm summer regions. 
Anonymized measurements are gathered from smart-meters collected by energy distributor Eneco at resolutions of 10 seconds over a period of 3 years starting from January 1\textsuperscript{st} 2019 to the 2\textsuperscript{nd} of August 2021. Weather data is assembled from publicly available Royal Netherlands Meteorological Institute (KNMI) weather stations measurements \cite{knmi}, that are paired to each building thanks to a geo-localization process using 4 (over the 6) ZIP code digits; an aggregation level that allows no anonymized user to be geographically isolated nor identified.

 Case Study 2 employs the open data set from the Building Data Genome project 2 (BDG2) \cite{miller2020building}. This open set was selected to allow reproducibility of our method while putting forward a benchmark for hierarchical forecasting in the building sector. The BDG2 includes 3053 energy meters from 1636 non-residential buildings grouped by site located in Europe and, principally, North America. The set covers two full years (2016–2017) at an hourly resolution with multi-meter building measurements paired with site weather data.

\subsection{Hierarchies}
\begin{figure}
    \centering
    \begin{adjustbox}{width=0.99\linewidth}
        \includegraphics{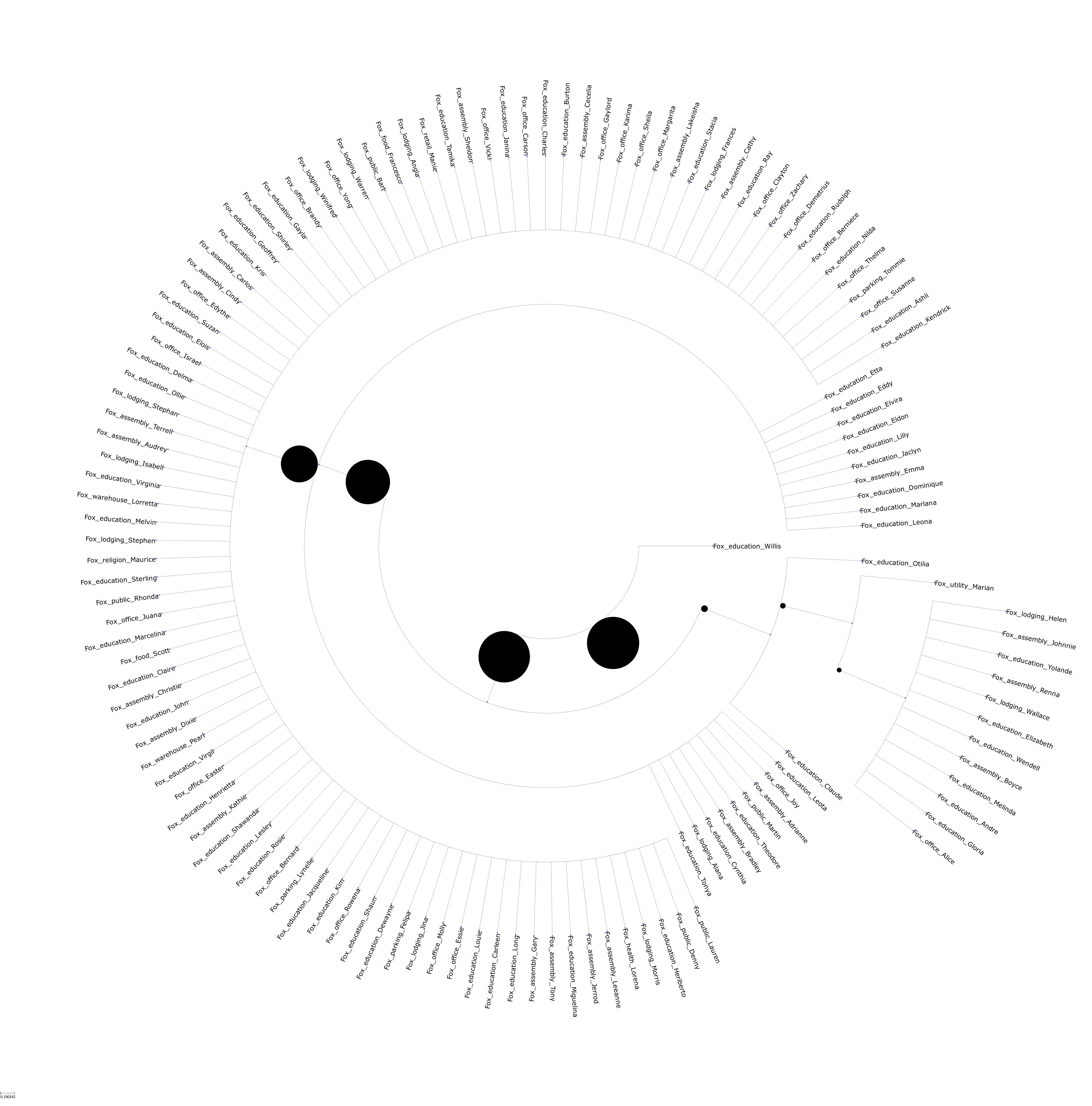}
    \end{adjustbox}
    \caption{Hierarchical spatial tree structure of the \textit{Fox} site from Case Study 2}
    \label{fig:foxtree}
\end{figure}
\begin{figure}
    \centering
    \begin{adjustbox}{width=0.80\linewidth}
        \includegraphics{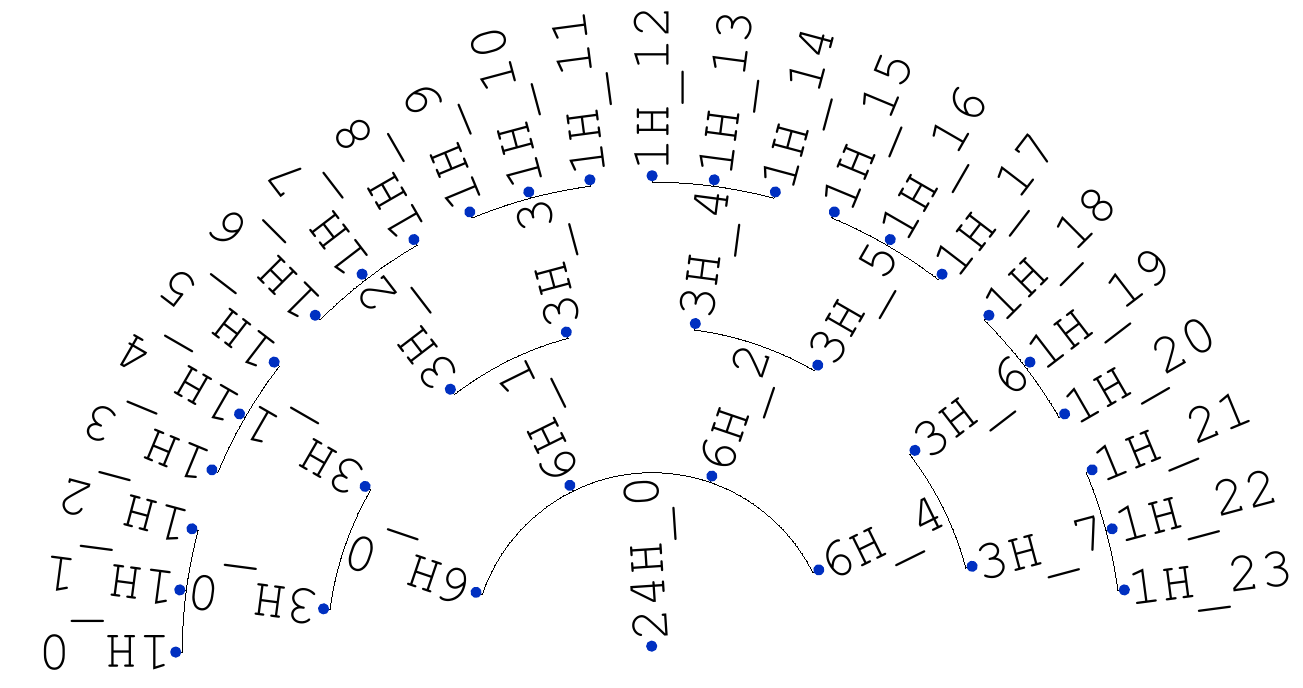}
    \end{adjustbox}
    \caption{Hierarchical temporal tree structure for day-ahead forecasts}
    \label{fig:temptree}
\end{figure}

\begin{center}
	\begin{table}
	\caption{Characteristics of assembled hierarchy per case study
	{\label{tab:hierarchy_charac}}}
	\vspace{-0.5em}
	\setlength\extrarowheight{-3pt}
	\begin{adjustbox}{width=0.96\linewidth}
	    \begin{tabular}{lllccc}
\toprule
  & \multicolumn{2}{l}{Characteristics}  &  Spatial & Temporal & Spatiotemporal \\
\midrule
  Case study 1 & n       & [\#]    & 383  & 37 & 14,171 \\
               & m       & [\#]    & 192  & 24 & 4,608 \\
               & horizon & [hours] & 1    & 24 & 24 \\
  \\
  Case study 2 & n       & [\#]    & 140  & 37 & 1,998 \\
               & m       & [\#]    & 133  & 24 & 1,200 \\
               & horizon & [hours] & 1    & 24 & 24 \\
\bottomrule
\end{tabular}
	\end{adjustbox}
	\end{table}
\end{center}

Spatial hierarchies are defined by hierarchically clustering the prediction target time series, i.e., electricity demand. This step is carried out employing the Ward variance minimization algorithm \cite{mullner2011modern}. The obtained hierarchy is reduced in size by \textit{cutting the tree} using a defined distance threshold over visual inspection of the derived dendrogram. In this way, hierarchical structures located below the defined distance threshold will be clustered together, effectively reducing the number of connection nodes of the tree. Figure \ref{fig:foxtree} illustrates the attained reduced tree of the \textit{Fox} site of case study 2.
Temporal hierarchies are considered with a horizon of one day (tree root sampling frequency) while reaching down to granularities of hourly sampling intervals (tree leaves). Aggregation levels encompass sampling frequencies every 6 and 3 hours, resulting in a tree with sampling frequencies of 1 day, 6 hours, 3 hours and 1 hour per k-level, as illustrated in Fig. \ref{fig:temptree}.
Spatio-temporal trees are then obtained as a result of the dimensional combination of spatial and temporal hierarchies, as detailed under Sec. \ref{subsec:multidim_S}. To limit the exponential explosion in tree size from dimensional combination, spatial trees are limited to 50 leaves in case study 2.
Table \ref{tab:hierarchy_charac} details the different characteristics of the considered hierarchies per case study.

\subsection{Model learning setup}\label{subsec:modelearningsetup}
In both case studies, we proceed to resample the time-series to hourly intervals. Time-series with no cumulative missing values larger than 2 hours are considered and smaller gaps are interpolated via a moving average using a window size of 8 hours.

\subsubsection{Feature engineering}
Data sets are then treated per dimensional batches, namely, per site, sub-site sample, or building for spatial, spatio-temporal, and temporal dimensional hierarchies respectively.
Features are selected based on their Maximum Information Coefficient (MIC) \cite{reshef2011detecting} computed in relation to the learning target. MIC is a powerful indicator that captured a wide range of associations both functional and not while providing a score that roughly equals the coefficient of determination (R2) of the data relative to the regression function. It ranges between values of 0 and 1, where 0 implies statistical independence and 1 a completely noiseless relationship. The advantage of using MIC for feature engineering over the more commonly employed person correlation indicator \cite{miller2019s} is that it captures non-linear relationships present in the data, which deep-learning models are popularly capable of detecting.  We retain features exhibiting MIC values higher than 0.25, as electric loads can typically become quite volatile and impede MIC values with noise.

Additionally, to feed the learner with the most relevant historical information of the predicted target, we select the 3 top auto-correlation values per temporal aggregation level above 0.25 as model input features. If no target auto-correlation value is above 0.25, we consider the most recent historical information, i.e., $t_k-1$ where $t_k$ is the first k-level time-step value of the predicted horizon.

Both MIC and autocorrelation selection thresholds are settings that should typically be included in the hyper-parameter optimization of the model validation phase. 
While evaluating the performance of hierarchical regressors over three varying dimensional considerations and two different case studies, this work considers the tuning of these thresholds to lay outside of its scope, as such computations rapidly become excessively burdensome.

\subsubsection{Data partitioning and transformation}\label{subsect:data_scaling}
\begin{figure}
    \centering
    \begin{adjustbox}{width=0.99\linewidth}
        \includegraphics{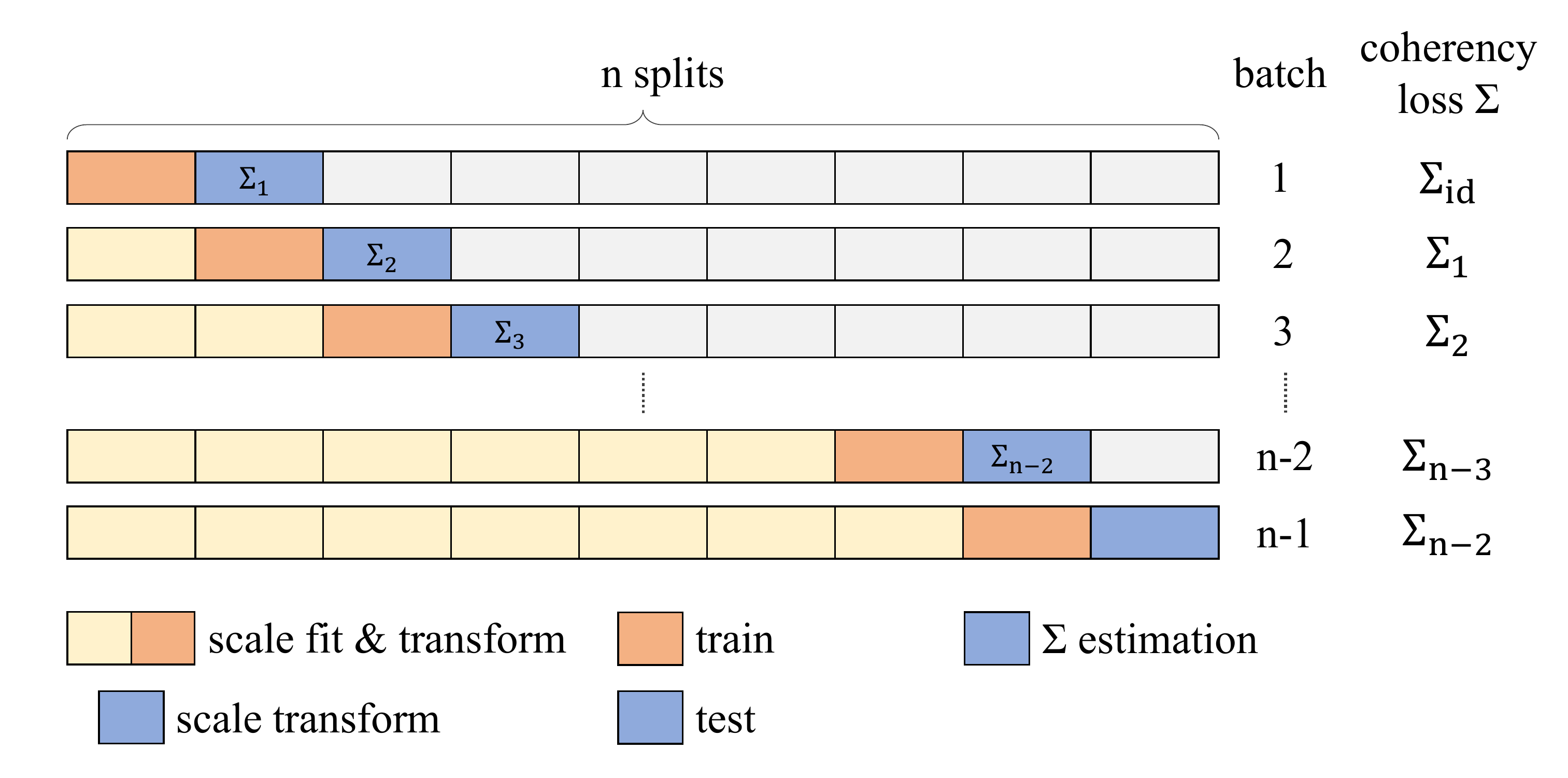}
    \end{adjustbox}
    \caption{Data partitioning, transformation and covariance matrix estimation setup}
    \label{fig:datasplit}
\end{figure}
Training and testing sets are then defined employing TimeSeriesSplit, a times-series cross-validator of the sklearn package \cite{scikit-learn}, with equal test-size in a rolling window setup. We next proceed to standard normalize the data per batch using batch-specific available historical information such that each batch-scaler is first fitted to the current, and past, training set. The fitted-scaler is then employed to transform batch-specific test sets, as depicted by Fig. \ref{fig:datasplit}.
This process avoids data leakage situations, which refers to the inadvertent use of data from test sets, or more generally data not available during inference while training a model. This typically occurs when the data is normalized prior to partitioning for cross-validation, i.e., by performing smoothing or normalization over the whole series before partitioning for training and testing \cite{hewamalage2022forecast}.

While it benefits the performance of (deep) neural networks to normalize input features and predicted target, i.e., from unscaled $\boldsymbol{y}_x$ to scaled $\boldsymbol{y}_z$, this shatters the hierarchical relationship of the regressors' outputs; thus, affecting the soundness of the coherency loss-function. 
To integrate the coherency loss function in such a setting, hierarchical relationships of predicted values $\hat{\boldsymbol{y}}_z$ are restored by reverse transformation prior to coherency loss calculation. The obtained reversed-scaled prediction $\hat{\boldsymbol{y}}_x$ is reconciled to $\tilde{\boldsymbol{y}}_x$ following Eq. \eqref{eq:sigma} and is finally re-scaled to $\tilde{\boldsymbol{y}}_z$ to calculate the coherency loss function against its original predicted self $\hat{\boldsymbol{y}}_z$.

\subsubsection{Coherency settings}
The estimation of the covariance matrix is performed over test sets. For the first batch training, we employ the identity covariance estimate $id$ as no forecasts are yet available. Each batch training $i$ then comes with a new covariance matrix estimate $\Sigma_i$ that is employed in the coherency loss function of the next training set $i+1$, see Fig. \ref{fig:datasplit}. This setup echoes the adaptive covariance matrix estimation proposed by \cite{bergsteinsson2021heat} employed for temporal hierarchies, anchored here quite organically in the learning process of neural networks.

\subsubsection{Designing hierarchical regressors}\label{subsubsec:modeldesign}
We select deep neural network regressors to best serve the benchmarking of hierarchical-coherent forecasts. Such machine-learning regressors possess well-developed packages supporting custom implementations that serve our approach well.
The regressor is structured as a series of sequential layers decreasing proportionally in size, from initial input layer size to the desired output dimension $n$. 
The optimal number of layers is selected from monitored test-set prediction performances while step-wise increasing the network's depths starting from shallow 1-layer perceptrons. This allows the selected architecture to serve an "as simple as possible yet as complex as necessary" design. 
Model hyper-parameters are later tuned over a concise grid encompassing loss function parameter $\alpha$, activation functions, and dropout fraction, further improving the performance of the model.
These tests resulted in the design of a deep neural network of 3 layers, leveraging sigmoid activation functions and dropout ratios of 0.2 on all but the last layer favoring a linear activation and no dropouts, and a retained $\alpha$ coefficient value of 0.75
The presented models of Sec. \ref{sec:hl} were implemented in Python using the Keras package \cite{chollet2015keras}.

    \section{Results}\label{sect_res}

We describe the outcome of the implementation here over spatial, temporal and spatio-temporal hierarchical structures per case study. In particular, we evaluate the accuracy and coherency of the forecasted building loads outlined in an annotated heatmap and bar plot respectively, where the presented coherency loss relates solely to the output of the forecasting method, i.e., reconciliation referred to \textit{None}, as reconciled forecasts all possess null coherency losses.
Evaluated forecasting methods cover the independent (\textit{base}), \textit{multi-task}, and \textit{hierarchical} forecasting methods presented under Sec. \ref{sec:hierarchical_forecasting}. Hierarchical forecasting and reconciliation methods each consider the covariance approximations presented under Fig. \ref{fig:covariance_method}, i.e., ordinary least square (\textit{id}), structural (\textit{str}), heterogeneous variance (\textit{hvar}), homogeneous variance (\textit{svar}), shrunken covariance (\textit{cov}) and shrunken k-level covariance (\textit{kcov}).
Necessary computational resources inherent to the forecasting methods are also discussed.

\subsection{Case study 1}

\subsubsection{Spatial}
\begin{figure}
    \centering
    \begin{adjustbox}{width=0.99\linewidth}
    \includegraphics{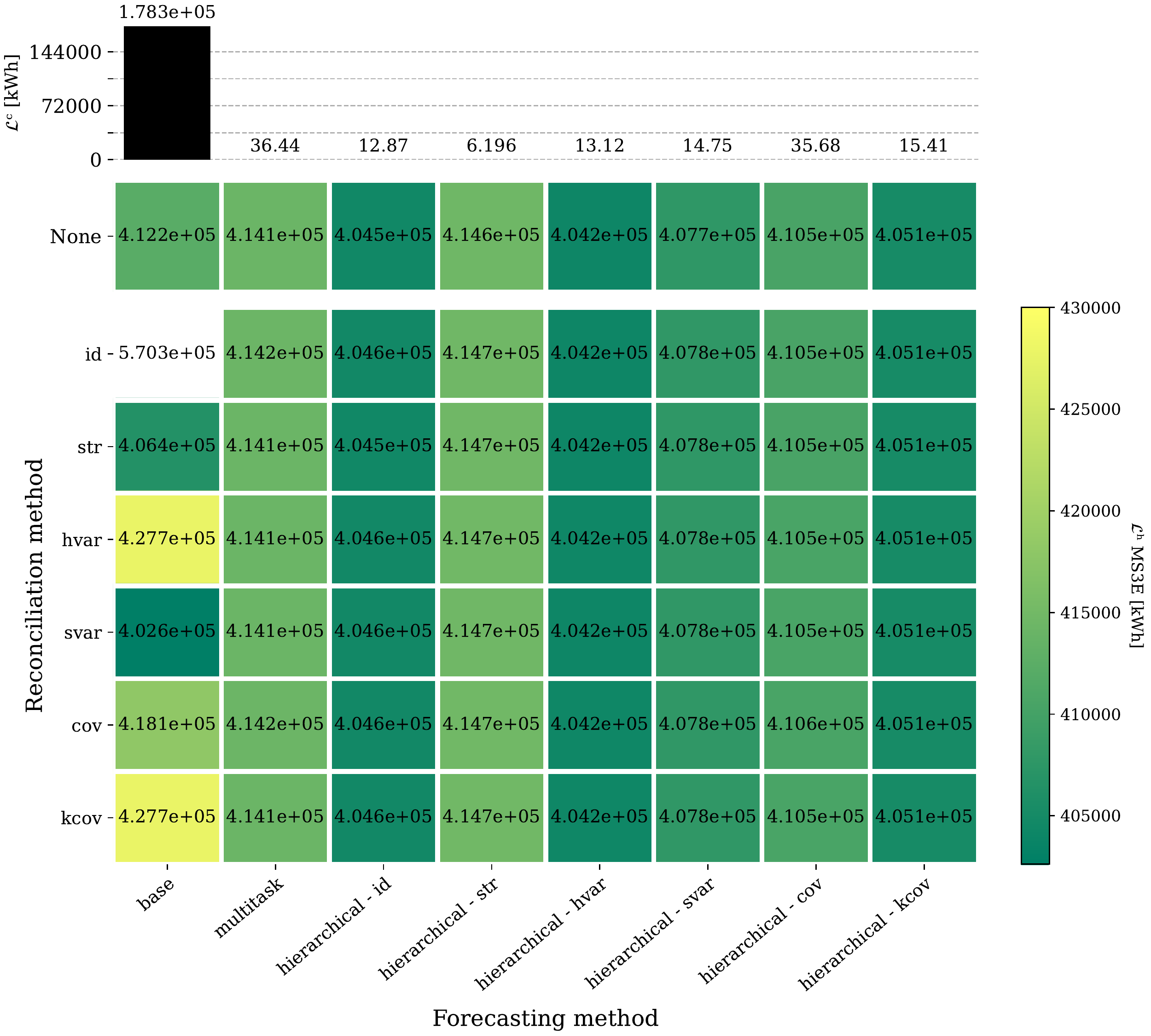}
    \end{adjustbox}
    \caption{Spatial hierarchy forecasting performance of case study 1. To allow the differentiation of performances across the heatmap, extreme values were cut off from the color map.}
    \label{fig:spatial_casestudy1}
\end{figure}
Performances of spatial hierarchical forecasts are presented under Fig. \ref{fig:spatial_casestudy1}, where illustrated hierarchical losses showcase \textit{svar} as the best hierarchical forecast performer, with and without reconciliation.
The lowest hierarchical MS3E originates from \textit{base} forecast reconciled with \textit{id} covariance matrix approximation, while \textit{hvar} and \textit{kcov} also notably perform quite poorly for this forecasting method.
Overall, the performance of the forecasts seems to rely more on the selected forecasting method rather than their reconciliation approaches.

Coherency losses seem in line with expected results; \textit{base} forecast is showcased as the most incoherent outcome, holding coherency errors ranging up to 1.783e5 kWh, while \textit{multi-task} and \textit{hierarchical} regressors score MS3Es of 36 kWh and 16 kWh (on average) respectively.

\subsubsection{Temporal}
\begin{figure}
    \centering
    \begin{adjustbox}{width=0.99\linewidth}
    \includegraphics{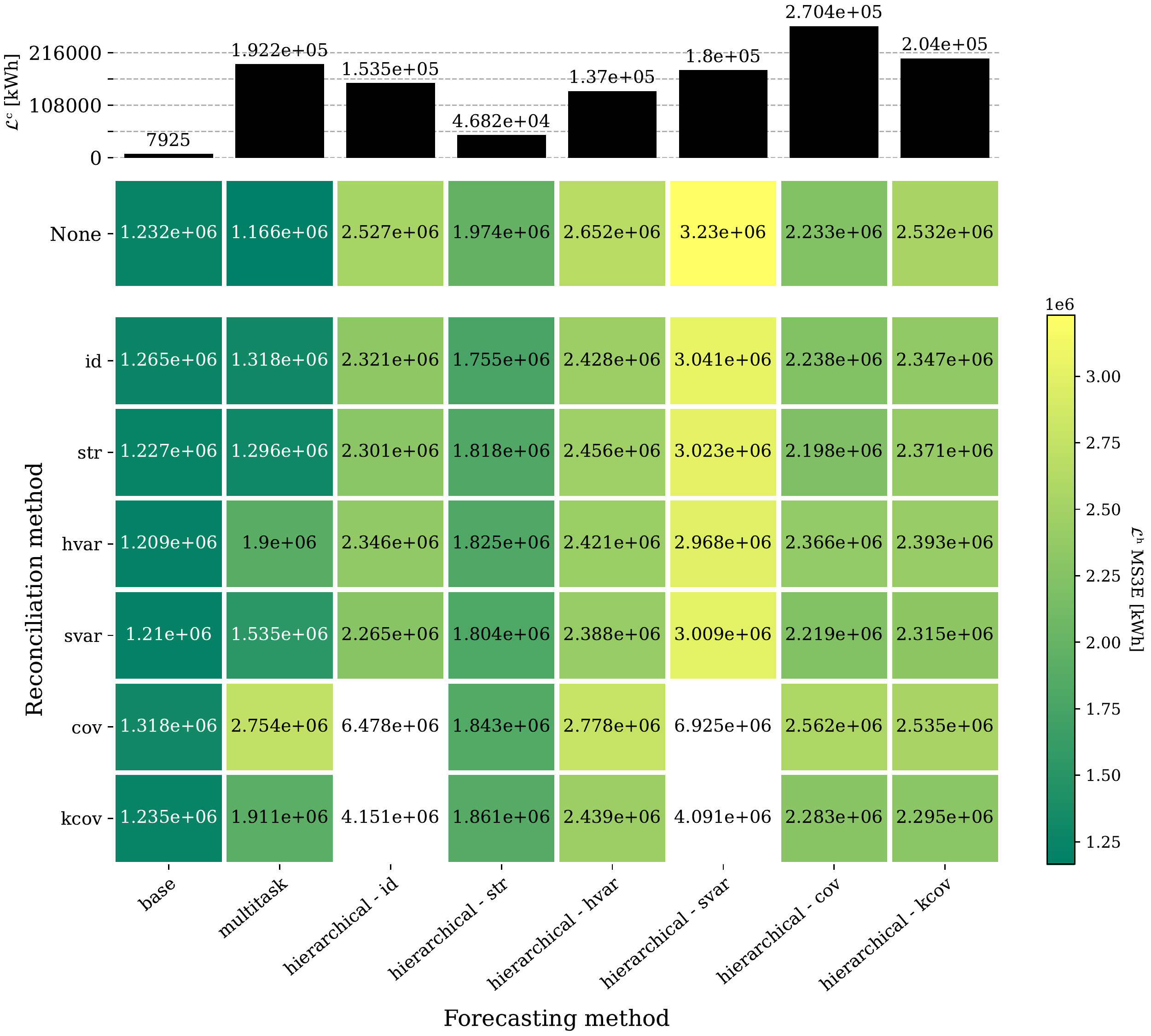}
    \end{adjustbox}
    \caption{Temporal hierarchy forecasting performance of 40 buildings from case study 1. To allow the differentiation of performances across the heatmap, extreme values were cut off from the color map.}
    \label{fig:temporal_casestudy1}
\end{figure}
Temporal hierarchical forecasting performances, on the other hand, portray a much different behavior. 
As illustrated by Fig. \ref{fig:temporal_casestudy1}, it is here the \textit{base} and \textit{multi-task} regressors that possess the lowest hierarchical losses, with 1.232e6 and 1.116e6 kWh respectively.
The poorer performer without reconciliation in this setup is \textit{svar}, with an MS3E of up to 3.23e6 kWh. Extreme poor performances are noticeable for the \textit{cov} and \textit{kcov} reconciliations of \textit{id} and \textit{svar} forecasting methods.
Overall, the performance of the forecasting methods here seems also more driven by the considered forecasting method than reconciliation.

In terms of coherency, the \textit{base} forecast surprisingly exhibits the most coherent outputs with an MS3E of 7925 kWh. It is followed by \textit{str}, and all other \textit{hierarchical} forecasts which compare considerably worst featuring inconsistency errors of 4.682e4 kWh and 1.691e5 kWh (on average) respectively.

\subsubsection{Spatio-temporal}
\begin{figure}
    \centering
    \begin{adjustbox}{width=0.99\linewidth}
    \includegraphics{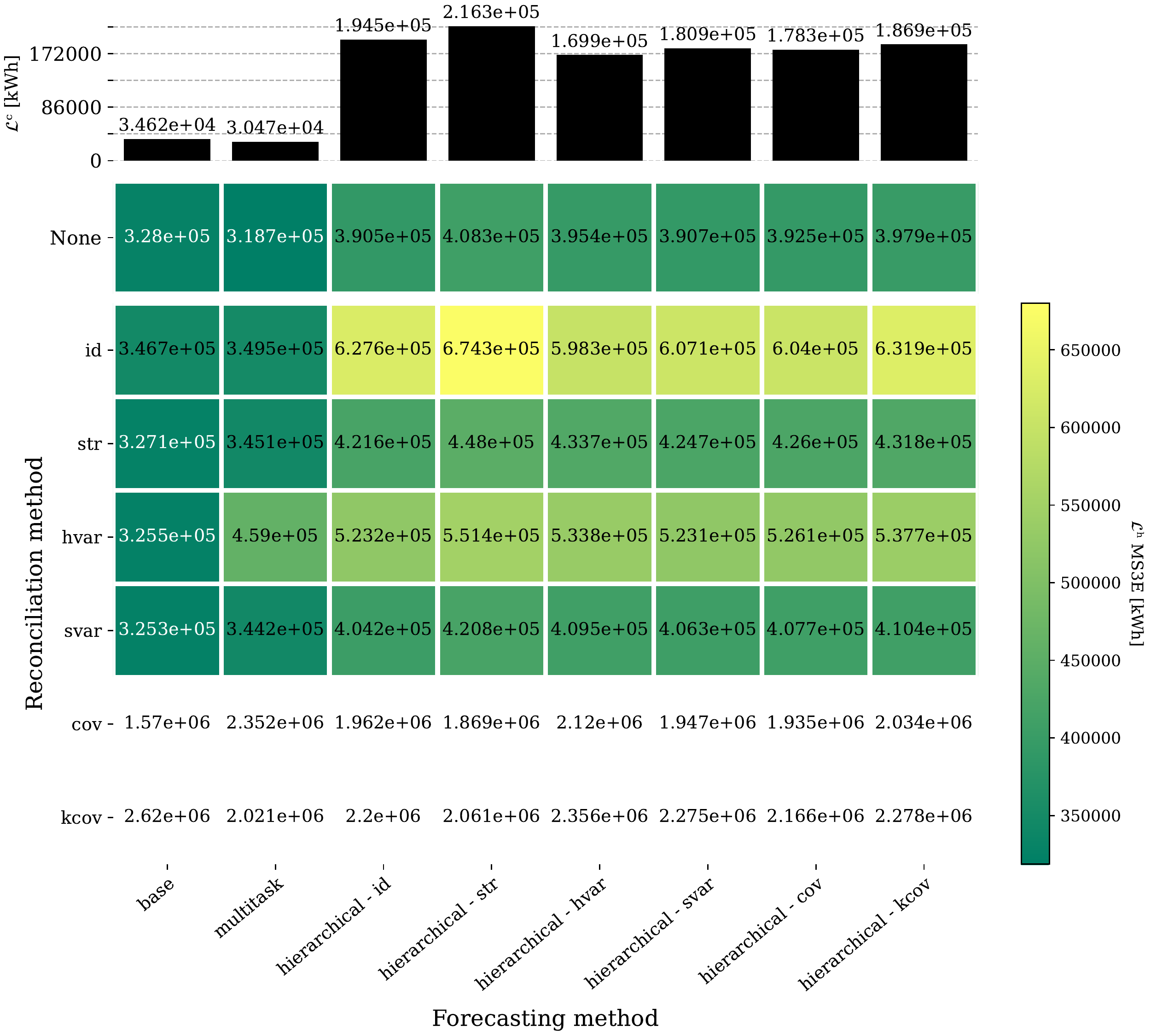}
    \end{adjustbox}
    \caption{Spatio-temporal hierarchy forecasting performance of 41 buildings from case study 1. To allow the differentiation of performances across the heatmap, extreme values were cut off from the color map.}
    \label{fig:spatiotemporal_casestudy1}
\end{figure}

Finally, spatio-temporal forecasting performances exposed in Fig. \ref{fig:spatiotemporal_casestudy1} reveal contrasting outcomes compared to previous hierarchies. 
First, all \textit{cov} and \textit{kcov} reconciliations here perform extremely poorly, irrespective of the forecasting method employed, with hierarchical losses ranging between 1.57e6 and 2.62e6 kWh.
Similarly to the temporal hierarchy, \textit{base} and \textit{multi-task} forecasts perform overall better than \textit{hierarchical} ones. The \textit{multi-task} regressor without reconciliation is showcased as the best performer in this setup with an MS3E of 3.187e5 kWh.
It can notably be observed here that all \textit{hierarchical} and \textit{multi-task} forecast reconciliations do not improve the accuracy of their original forecast. 
Additionally, exposed performances here display a much stronger dependency on the considered reconciliation approach than forecasting.

Concerning coherency losses, spatio-temporal hierarchies produce two distinct performances; where \textit{base} and \textit{multi-task} forecasts exhibit inconsistencies of 1 order of magnitude lower than all \textit{hierarchical} ones, i.e., 3.255e4 kWh against 1.878e5 kWh on average.

\subsection{Case study 2}
\begin{figure}
    \centering
    \begin{adjustbox}{width=0.99\linewidth}
        \includegraphics{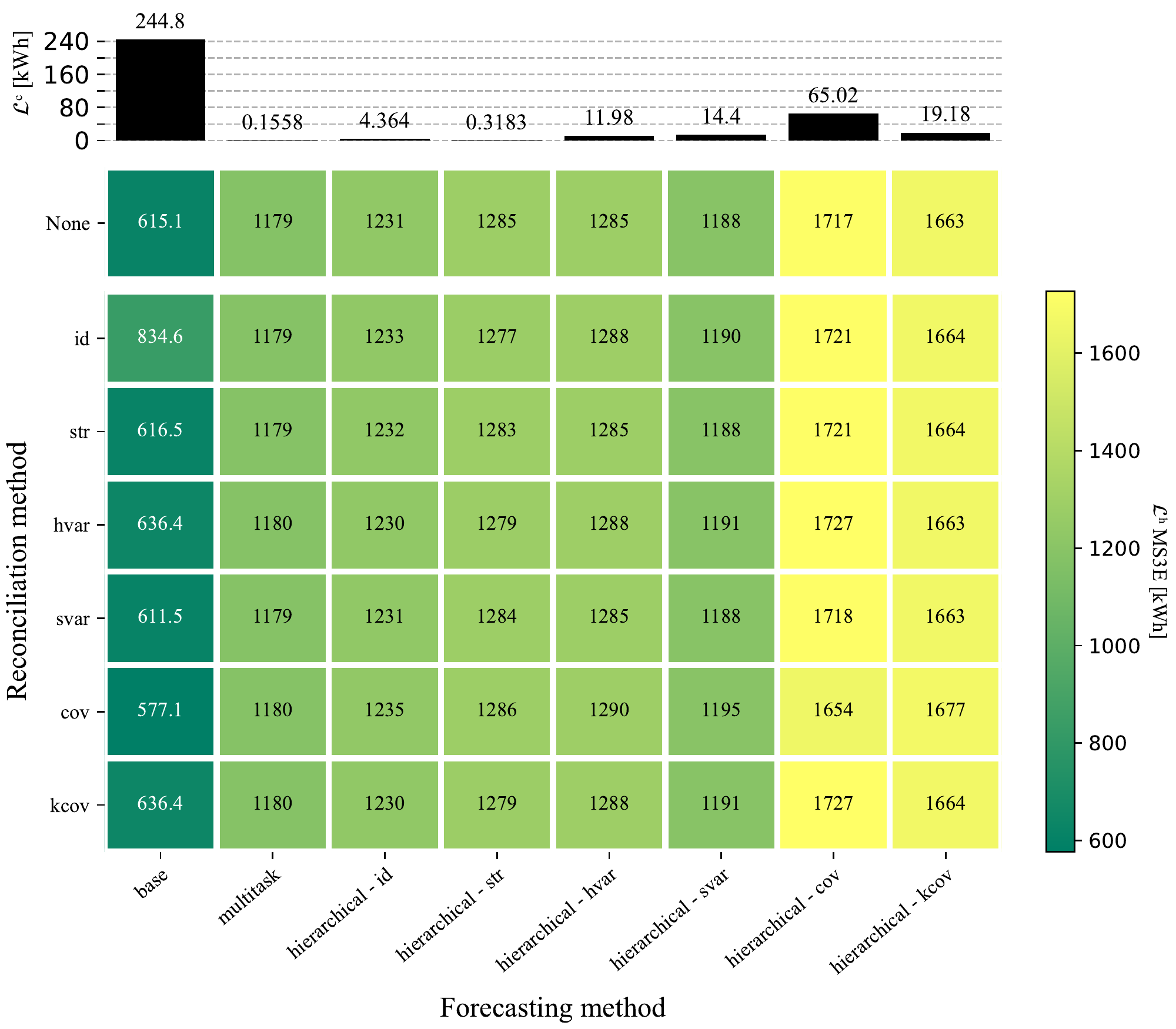}
    \end{adjustbox}
    \caption{Spatial hierarchy forecasting performance of the \textit{Fox} site of case study 2}
    \label{fig:spatial_casestudy2}
\end{figure}

Concerning case study 2, the spatial hierarchical forecasting performance presented under Fig. \ref{fig:spatial_casestudy2}, depicts noticeable variations from case study 1.

\subsubsection{Spatial}

Here, the \textit{base} case exhibits the most accurate forecast, although at the cost of higher inconsistencies across the tree.
\textit{Multi-task} forecasts followed by structural, \textit{str}, hierarchical ones both produce the most coherent outcomes. Surprisingly, while the \textit{multi-task} forecast is trained without coherency information, its forecast displays the best coherency performance in this scenario.

Overall, the best forecast accuracy is obtained from \textit{base} forecasting reconciled with the \textit{cov} approximation, while the worst performer for this scenario is the \textit{cov} hierarchical forecasting with either \textit{kcov} or \textit{hvar} covariance approximations. It displays hierarchical MS3Es ranging from 611.5 to 1.727e3 kWh and coherency MS3Es varying between 0.156 and 245 kWh.

\subsubsection{Temporal}
\begin{figure}
    \centering
    \begin{adjustbox}{width=0.99\linewidth}
        \includegraphics{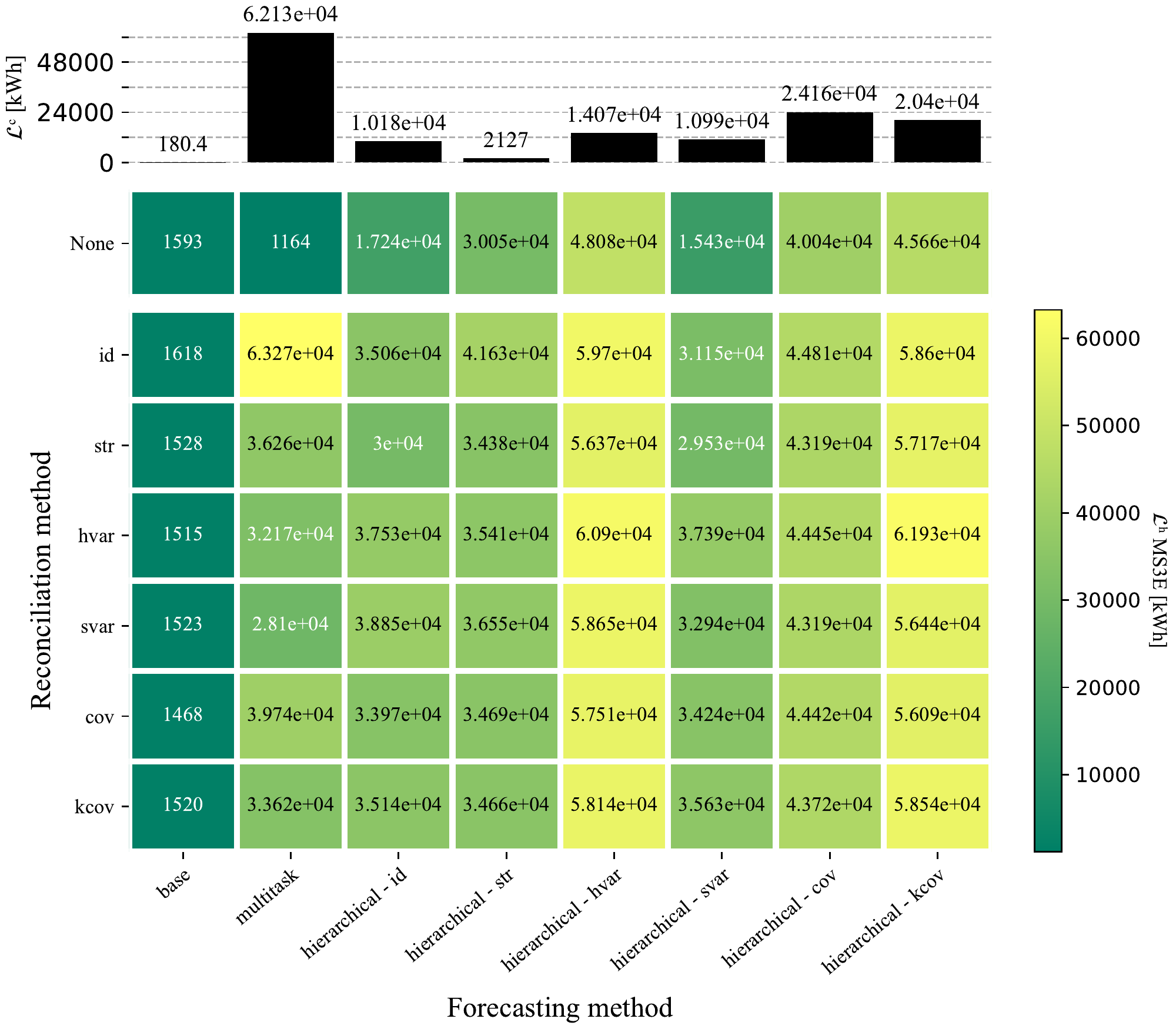}
    \end{adjustbox}
    \caption{Temporal hierarchy forecasting performance of 66 buildings from the \textit{Fox} site of case study 2}
    \label{fig:temporal_casestudy2}
\end{figure}

The averaged temporal hierarchical forecast performance of 66 buildings from the \textit{Fox} site is exposed under Fig. \ref{fig:temporal_casestudy2}.
Forecasting performances are overall significantly worse than those of spatial-hierarchies, with hierarchical MS3Es now ranging between 1.164e3 and 6.327e4 kWh, while coherency losses fluctuate from 180 to 6.213e4 kWh; an order of magnitude about 3 times higher than temporal trees. 
Here, the best-performing forecast belongs to the \textit{multi-task} forecast with no reconciliation, which also displays the highest inconsistency score.
The lowest performing forecast interestingly resides with the \textit{id} reconciliation of that same forecast.

The most coherent forecast produced for temporal-trees peculiarly originate from \textit{base} forecasts, which neither share information across the hierarchy, nor possess coherency-knowledge. Other \textit{hierarchical} forecasts produce coherency losses ranging between 2.120e3 and 2.416e4 kWh.

\subsubsection{Spatio-temporal}
\begin{figure}
    \centering
    \begin{adjustbox}{width=0.99\linewidth}
        \includegraphics{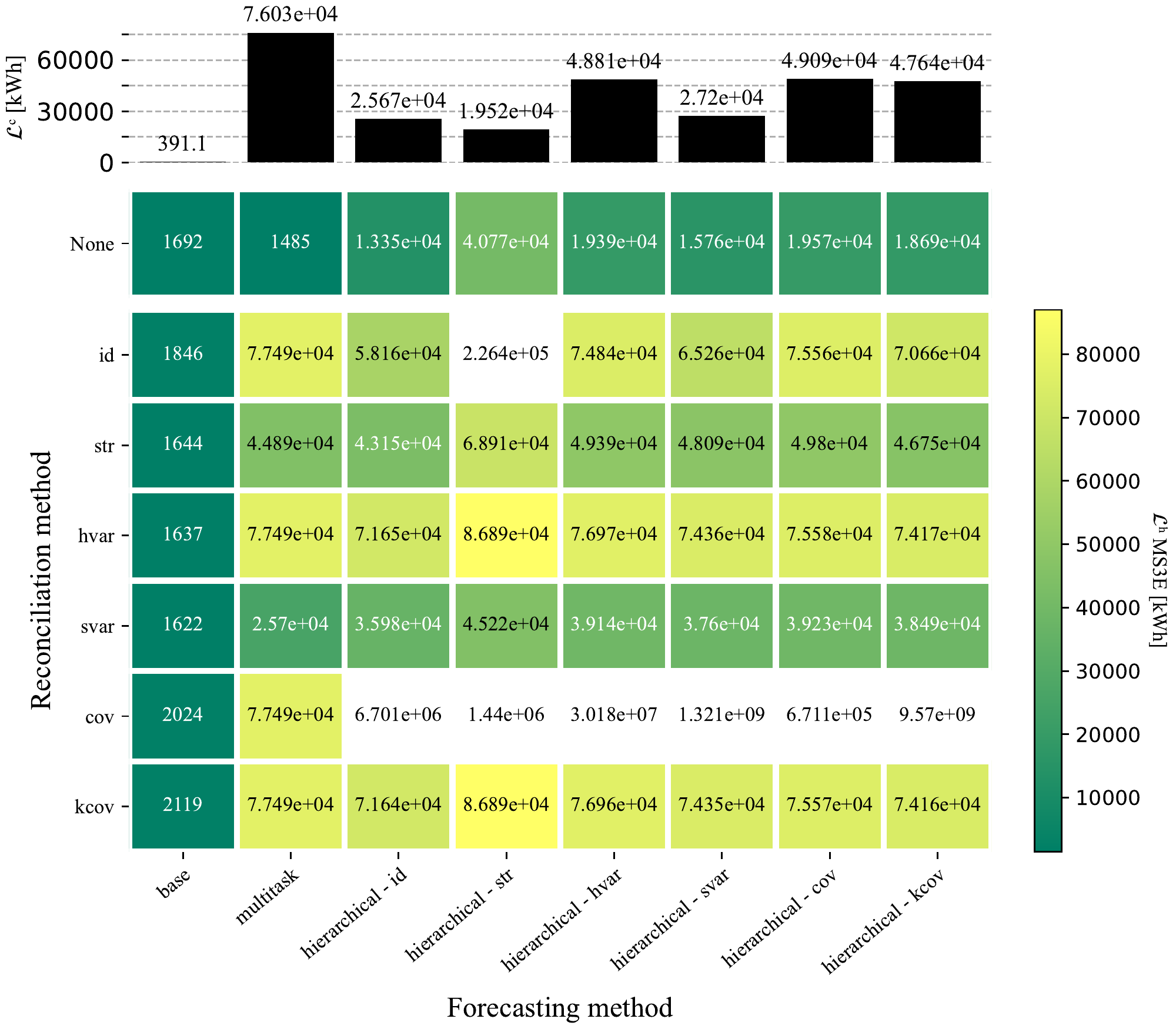}
    \end{adjustbox}
    \caption{Spatio-temporal hierarchy forecasting performance of 50 buildings from the \textit{Fox} site of case study 2. To allow the differentiation of performances across the heatmap, extreme values were cut off from the color map.}
    \label{fig:spatiotemporal_casestudy2}
\end{figure}

Lastly, the forecast performance of spatio-temporal structures considering 50 buildings from the \textit{Fox} site is presented under Fig. \ref{fig:spatiotemporal_casestudy2}.
Similarly to the temporal-tree, hierarchical losses display much poorer performances compared to their spatial antecedent, with MS3Es ranging between 1.485e3 and extreme 9.57e9 kWh values, while coherency losses vary from 391 to 7.603e4 kWh.

Mirroring the results from temporal-hierarchies, the forecasting technique withholding the lowest hierarchical loss is the \textit{multi-task} learner without reconciliation which is also characterized by the highest coherency loss.
A series of extreme poor performers are identified as a result of the \textit{cov} reconciliation over all hierarchical-learners.
Contrary to temporal-tree, reconciled forecasts performances here seem driven by the reconciliation method rather than the considered forecasting technique.

Coherency scores display overall poor performances across all \textit{hierarchical} and \textit{multi-task} learners with losses ranging 2 orders of magnitude higher than the best case \textit{base} regressor.

\subsection{Computational prospects}
\begin{table}
 \begin{center}
    \caption{Averaged computing times (in seconds) of evaluated forecasting methods{\label{tab:computation}}}
    \vspace{-0.5em}
    \setlength\extrarowheight{-3pt}
    \begin{adjustbox}{width=0.85\linewidth}
        \begin{tabular}{lc|ccc}
\toprule
& tree size & \multicolumn{3}{c}{forecasting method} \\
& n &  base  & multi-task & hierarchical \\
\midrule
Case study 1 & 14,171 & 3.6 & 392 & 397.3 \\
             & 383 & 34.9 & 90 & 96.6 \\
             & 37 & 12.2 & 12 & 13 \\
Case study 2 & 1,998 & 2.5 & 70 & 77 \\
             & 140 & 22.5 & 70 & 80.6 \\
             & 37 & 2.2 & 20 & 19.3 \\
\bottomrule
\end{tabular}

    \end{adjustbox}
 \end{center}
 \vspace{-0.5cm}
\end{table}

Computational performances of forecasting approaches are here considered, providing a complete overview of evaluated methods.
Table \ref{tab:computation} presents the computation time of each forecasting method averaged over all training batches.
Two anticipated findings can be noted from it. 

First, the computing time is positively correlated to the size of the hierarchy. One exception seems to deviate from that rule in case study 2, between tree sizes of 1,998 and 140, which display relatively close computing times. 
Second, smaller regressors, i.e, \textit{base}, train faster than larger ones, namely, \textit{multi-task} and \textit{hierarchical}.
Both these observations can be explained by the increasing number of weights to update in the larger regressor. The more weights to update, the longer the training will take.

Although independent regressors seem attractive due to their noticeably faster computing times, it should be noted that the displayed performances depict only the average computing time of a unique independent regressor. Should such regressors not be trained and tested in a distributed computational setup, then these numbers would need to be multiplied by the hierarchy size to obtain an appropriate estimation of the required computing period.

    \section{Discussion}\label{sect_app}
Although presented case studies bear varying results, these also display a number of commonalities supporting interpretation and analysis, which are here discussed.

\subsection{Hierarchical-coherency value}
Unifying the forecast of hierarchical structures under one regressor possesses attractive data-efficient prospects, i.e., cross-tree information exchange combined with embedded-structural learning provided from coherency loss. However, produced outcomes from hierarchical-coherent learners were only found to bring added value in one setting, namely, the spatial hierarchy of case study 1.
This can be explained by the similarities in building loads of case study 1, which encompassed time series of similar patterns and dynamics, all originating from residential constructions, while case study 2 included a broader collection of construction types covering offices, college classrooms, lodging, warehouses, and parkings. 
Such profile diversities are challenging to learn from limited measurements, particularly for a large model involving considerable numbers of regression weights.

It can consequently be found that while the results of the spatial hierarchy of case study 1 are promising, these unveil, in fact, important challenges hierarchical forecasting must face.


\subsection{An efficient but arduous learning process}
    Although the outcome of hierarchical learning demonstrated promising performances, identified in the spatial hierarchy of case study 1, the resulting number of weights to update and possibly conflicting forecasted outputs can become burdensome, i.e., as unveiled by the performance of the spatial hierarchy of case study 2.
    Indeed, with hierarchical regressors growing in size, their number of neuron connections increases by an exponential factor of 2. This renders the learning process of these models laborious as more data should support the learning of larger number of weights.
    Additionally, multi-output regressors are faced with the challenging task of predicting numerous outcomes which might exhibit highly different, possibly antipodal, dynamics. This also affects the learning process, which might struggle to identify these discrepancies from limited training data.
    
\subsection{Induced coherency over accuracy}
    Overall, temporal hierarchies of the considered case studies were seen to perform significantly worse than spatial ones. This significant change can be attributed to the combination of two factors.
    First, the longer forecasting horizon of temporal trees compared to spatial ones, i.e., 24 hours against 1, implies that forecasts must rely on fewer data and less recent information while dealing with higher uncertainties, thus negatively affecting their performances.
    Secondly, building electrical loads are endowed with a periodicity that falls precisely on the forecasted horizon of 24 hours. This consequently leads to little variations in the forecasted element of its hierarchy. And, while this characteristic is desirable for ordinary forecasting, the addition of the coherency-loss function, although weighted by the $\alpha$ coefficient - see Eq. \eqref{eq:Lhc}, may push the regressor to produce constant predictions, tailored more to coherency than accuracy, thus resulting in unrealistic and inaccurate predictions.
    
\subsection{Faulty coherent-learning from normalized trees}
\begin{figure}
    \centering
    \begin{adjustbox}{width=0.75\linewidth}
        \includegraphics{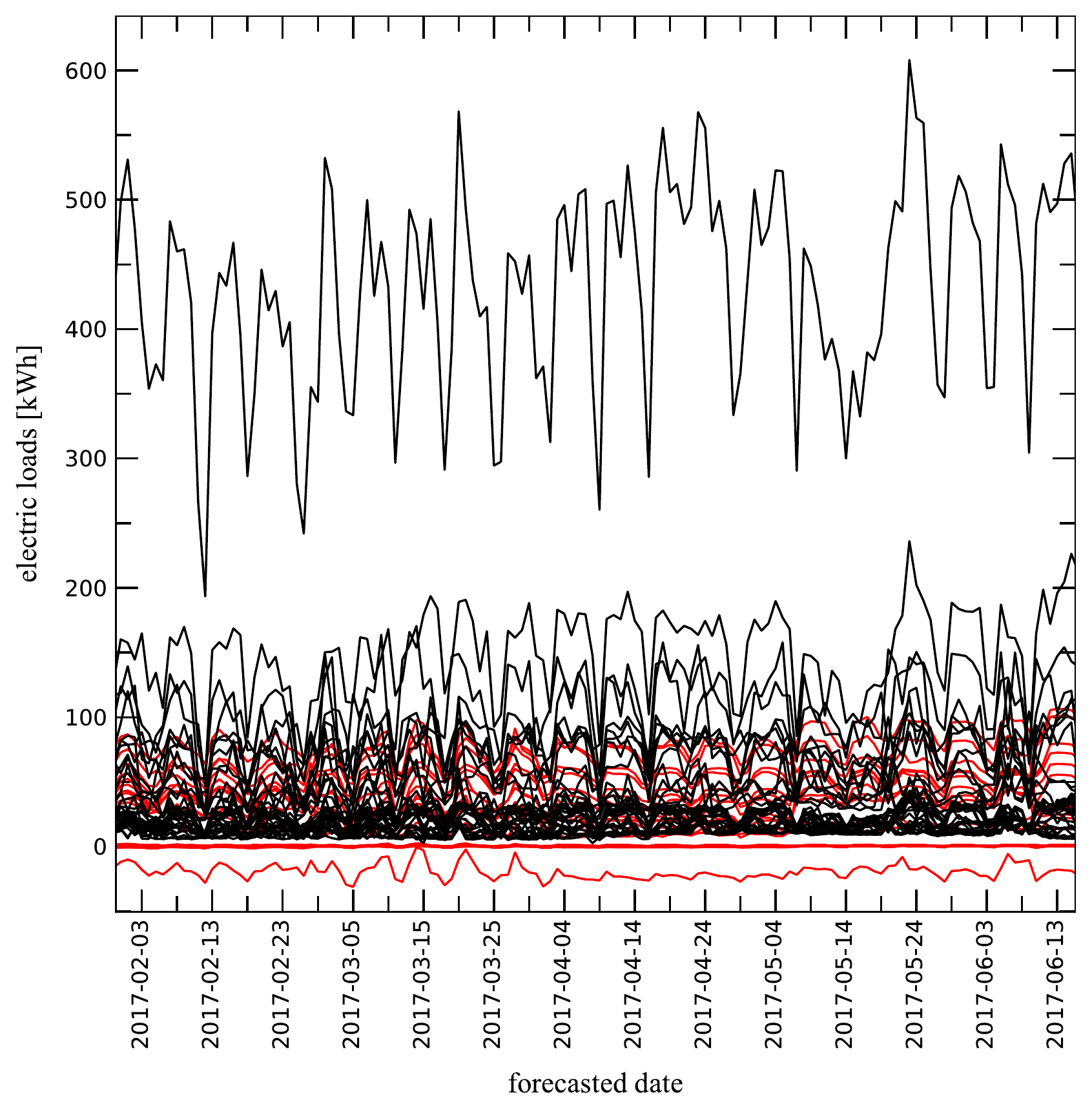}
    \end{adjustbox}
    \caption{Illustration of faulty coherent-learning from normalized trees. The predicted (red) versus true (black) electric loads of the \textit{Fox\_assembly\_Lakeisha} temporal hierarchical tree showcase the mirrored top-level forecast predicted in the negative domain.}
    \label{fig:faultycoherency}
\end{figure}

    In some settings, hierarchical-coherent learning displayed particularly poor performances from extreme hierarchical and coherency errors, i.e., temporal and spatio-temporal hierarchies.
    Following further inspection, it was noticed that these poor performers all withheld abnormal top-level forecasts which mirrored their expected true values in the negative domain, as illustrated in Fig. \ref{fig:faultycoherency}.
    These undesirable, yet peculiarly common, results can be traced back to the normalization of the target hierarchical time series. Indeed, while neural networks benefit from normalized targets, serving fair and balanced learning across the network's weights, this also shatters the coherency structure of the tree.
    The existing setup, detailed in Sec. \ref{subsect:data_scaling}, proceeds to tackle this issue by reverse-transforming these target values prior to the coherency constraint computation and re-scaling them for coherency loss calculation. This ensures both loss functions, namely hierarchical and coherency, see Eqs. \eqref{eq:Lh} and \eqref{eq:Lc} respectively, to operate on akin normalized time series. However, coherency learning can eventually produce adjustments larger than the original normalization ranges, e.g., lowering the top-level forecast $\hat{\boldsymbol{y}}_z$ fully into the negative domain such that the reverse standard transformation $\hat{\boldsymbol{y}}_x = \hat{\boldsymbol{y}}_z \cdot u + s$, where $u$ and $s$ refer to the mean and standard deviation of the fitted time series respectively, also produces a fully negative reverse-scaled $\hat{\boldsymbol{y}}_x$. 
    This evidently improper outcome consequently negatively impacts both the learning and the forecasting performance of the regressor and should be dealt with in future work.

    \section{Conclusion}\label{sect_con}
Ensuring coherent previsions of the future is crucial to support better informed and aligned decision-making processes across hierarchical structures.
And while previous works have attempted to exploit spatio-temporal hierarchical reconciliation using disparate steps \cite{kourentzes2019cross,spiliotis2020cross,punia2020cross,di2021cross}, no common formulation of multi-dimensional hierarchical structures had, to this date, been proposed. 
Furthermore, traditional hierarchical forecasts use disjointed forecasting and reconciliation processes that inherently deprive forecasting algorithms of (\textit{i}) the benefits of information transfer across (hierarchical) models, as well as (\textit{ii}) capitalizing on the coherency requirements of the produced forecast.
This paper proposes a solution to these shortcomings.

First, by formally defining multi-dimensional hierarchical structures, it extends conventional hierarchical forecasting methods, allowing the exploitation of spatio-temporal structures unified under a common frame, i.e., a unique summation and covariance matrix resulting from spatio-temporal function composition.

Second, rather than considering reconciliation a posteriori to forecasting, this work brings together independent forecasting models into a unique machine-learning regressor embedded with coherency information. 
This provides the regressor with (\textit{i}) a global overview of information across its hierarchy, permitting a cross-dimensional and data-rich learning process, while (\textit{ii}) learning coherency-requirements as a soft constraint thanks to a custom hierarchical-coherent loss function.
The approach can notably be tuned thanks to an adjustable $\alpha$ coefficient to either consider multi-task, hierarchical or only reconciliation in its learning process. 
Coherency of the produced hierarchical forecasts can then be enforced as a hard constraint using established reconciliation technics.
The outcome is a unified and coherent forecast across all examined dimensions, granting a common view of the future serving aligned and better decision-making.
The approach provides a data-driven solution to assemble diverging parts of an organization and blend information from varying sources, hierarchy levels, or scales \cite{kourentzes2019cross}.

Third, we evaluated our approach on two different case studies, across all hierarchical dimensions, considering established state-of-the-art reconciliation approaches. 
Results revealed spatial hierarchies to perform best while temporal and spatiotemporal structures suffered from coinciding forecasted horizon with the periodicity of electric loads from buildings.
Although the value potential of hierarchical-coherent learning was observed in case study 1, the performances of the approach were quite disparate in other settings. 
In this regard, a comprehensive analysis was reported revealing important challenges the approach faces. In particular, dealing with predicted outputs of conflicting trends while fitting an exponentially large number of weights to the model is a recurring fragility of the approach. Additionally, correcting faulty coherency training from normalized tree structures is another frailty future work may tackle.

Finally, to encourage knowledge dissemination we render our work fully replicable by open-sourcing developed python implementations under the public GitHub repository \href{https://github.com/JulienLeprince/hierarchicallearning}{https://github.com/JulienLeprince/hierarchicallearning}.

\subsection{Outlooks and future work}
This paper proposes a novel hierarchical learning method yielding important implications for forecasting theory. Indeed, by directly forecasting hierarchies this work opens the door to leveraging multi-scale and multi-frequency measurement information driving improved forecast accuracies. It notably expands and unites traditionally disjointed methods together providing a path toward a novel generation of forecasting regressors.

Meanwhile, numerous directions for future work can already be distinguished, guiding attempts to tackle uncovered obstacles.
As such, the curse of dimensionality endowed from larger, unified hierarchical models can notably be undertaken by investigating distributed and connected models working as a hybrid solution between independent, but tractable regressors and extensive hierarchical ones.
In addition, varying native multi-output machine learning algorithms may be examined such as ensemble decision trees, Gaussian processes, K-Neighbors, long short-term memory neural networks, and support vector machines. In particular, algorithms that deal with different ranges of target values can naturally tackle issues with coherency learning due to scaling.
Finally, comparing hierarchical learning performances against established models, i.e., grey- or white-box, that benefit from the inclusion of domain expertise to tackle targeted behaviors, such as seasonality, advances another interesting endeavor for future work.

\section{CRediT authorship contribution statement}\label{sect_credit}

\textbf{Julien Leprince}: Conceptualization, Methodology, Software, Formal analysis, Investigation, Data curation, Writing - original draft, Writing - review and editing, Visualization.
\textbf{Henrik Madsen}: Methodology, Supervision, Validation, Writing - review and editing.
\textbf{Jan Kloppenborg M{\o}ller}: Methodology, Supervision, Validation, Writing - review and editing.
\textbf{Wim Zeiler}: Supervision, Funding acquisition.

All authors have read and agreed to the published version of the manuscript.

\section{Acknowledgments}
This work is funded by the Dutch Research Council (NWO), in the context of the call for Energy System Integration \& Big Data (ESI-bida).
We gratefully acknowledge the support and contribution of Eneco with particular thanks to Dr. Kaustav Basu and Rik van der Vlist for this research, as well as SEM4Cities, funded by Innovation Fund Denmark (Project No. 0143-0004).

\bibliography{main}

\begin{thebibliography}{10}
\expandafter\ifx\csname url\endcsname\relax
  \def\url#1{\texttt{#1}}\fi
\expandafter\ifx\csname urlprefix\endcsname\relax\def\urlprefix{URL }\fi
\expandafter\ifx\csname href\endcsname\relax
  \def\href#1#2{#2} \def\path#1{#1}\fi

\bibitem{nystrup2020temporal}
P.~Nystrup, E.~Lindstr{\o}m, P.~Pinson, H.~Madsen, Temporal hierarchies with
  autocorrelation for load forecasting, European Journal of Operational
  Research 280~(3) (2020) 876--888.

\bibitem{kourentzes2019cross}
N.~Kourentzes, G.~Athanasopoulos, Cross-temporal coherent forecasts for
  australian tourism, Annals of Tourism Research 75 (2019) 393--409.

\bibitem{athanasopoulos2009hierarchical}
G.~Athanasopoulos, R.~A. Ahmed, R.~J. Hyndman, Hierarchical forecasts for
  australian domestic tourism, International Journal of Forecasting 25~(1)
  (2009) 146--166.

\bibitem{kremer2016sum}
M.~Kremer, E.~Siemsen, D.~J. Thomas, The sum and its parts: Judgmental
  hierarchical forecasting, Management Science 62~(9) (2016) 2745--2764.

\bibitem{spiliotis2021hierarchical}
E.~Spiliotis, M.~Abolghasemi, R.~J. Hyndman, F.~Petropoulos, V.~Assimakopoulos,
  Hierarchical forecast reconciliation with machine learning, Applied Soft
  Computing 112 (2021) 107756.

\bibitem{taieb2021hierarchical}
S.~B. Taieb, J.~W. Taylor, R.~J. Hyndman, Hierarchical probabilistic
  forecasting of electricity demand with smart meter data, Journal of the
  American Statistical Association 116~(533) (2021) 27--43.

\bibitem{ahmad2020review}
T.~Ahmad, H.~Zhang, B.~Yan, A review on renewable energy and electricity
  requirement forecasting models for smart grid and buildings, Sustainable
  Cities and Society 55 (2020) 102052.

\bibitem{peng2021flexible}
J.~Peng, A.~Kimmig, Z.~Niu, J.~Wang, X.~Liu, J.~Ovtcharova, A flexible
  potential-flow model based high resolution spatiotemporal energy demand
  forecasting framework, Applied Energy 299 (2021) 117321.

\bibitem{spiliotis2020cross}
E.~Spiliotis, F.~Petropoulos, N.~Kourentzes, V.~Assimakopoulos, Cross-temporal
  aggregation: Improving the forecast accuracy of hierarchical electricity
  consumption, Applied Energy 261 (2020) 114339.

\bibitem{stone1942precision}
R.~Stone, D.~G. Champernowne, J.~E. Meade, The precision of national income
  estimates, The Review of Economic Studies 9~(2) (1942) 111--125.

\bibitem{weale1988reconciliation}
M.~Weale, The reconciliation of values, volumes and prices in the national
  accounts, Journal of the Royal Statistical Society: Series A (Statistics in
  Society) 151~(1) (1988) 211--221.

\bibitem{hyndman2011optimal}
R.~J. Hyndman, R.~A. Ahmed, G.~Athanasopoulos, H.~L. Shang, Optimal combination
  forecasts for hierarchical time series, Computational statistics \& data
  analysis 55~(9) (2011) 2579--2589.

\bibitem{wickramasuriya2019optimal}
S.~L. Wickramasuriya, G.~Athanasopoulos, R.~J. Hyndman, Optimal forecast
  reconciliation for hierarchical and grouped time series through trace
  minimization, Journal of the American Statistical Association 114~(526)
  (2019) 804--819.

\bibitem{10.1007/978-3-319-18732-7_15}
T.~van Erven, J.~Cugliari, Game-theoretically optimal reconciliation of
  contemporaneous hierarchical time series forecasts, in: A.~Antoniadis, J.-M.
  Poggi, X.~Brossat (Eds.), Modeling and Stochastic Learning for Forecasting in
  High Dimensions, Springer International Publishing, Cham, 2015, pp. 297--317.

\bibitem{TIMMERMANN2006135}
A.~Timmermann,
  \href{https://www.sciencedirect.com/science/article/pii/S1574070605010049}{Chapter
  4 forecast combinations}, Vol.~1 of Handbook of Economic Forecasting,
  Elsevier, 2006, pp. 135--196.
\newblock \href {https://doi.org/https://doi.org/10.1016/S1574-0706(05)01004-9}
  {\path{doi:https://doi.org/10.1016/S1574-0706(05)01004-9}}.
\newline\urlprefix\url{https://www.sciencedirect.com/science/article/pii/S1574070605010049}

\bibitem{8453006}
Y.~Zhang, J.~Dong, Least squares-based optimal reconciliation method for
  hierarchical forecasts of wind power generation, IEEE Transactions on Power
  Systems (2018) 1--1\href {https://doi.org/10.1109/TPWRS.2018.2868175}
  {\path{doi:10.1109/TPWRS.2018.2868175}}.

\bibitem{ATHANASOPOULOS201760}
G.~Athanasopoulos, R.~J. Hyndman, N.~Kourentzes, F.~Petropoulos,
  \href{https://www.sciencedirect.com/science/article/pii/S0377221717301911}{Forecasting
  with temporal hierarchies}, European Journal of Operational Research 262~(1)
  (2017) 60--74.
\newblock \href {https://doi.org/https://doi.org/10.1016/j.ejor.2017.02.046}
  {\path{doi:https://doi.org/10.1016/j.ejor.2017.02.046}}.
\newline\urlprefix\url{https://www.sciencedirect.com/science/article/pii/S0377221717301911}

\bibitem{spiliotis2019improving}
E.~Spiliotis, F.~Petropoulos, V.~Assimakopoulos, Improving the forecasting
  performance of temporal hierarchies, Plos one 14~(10) (2019) e0223422.

\bibitem{bergsteinsson2021heat}
H.~G. Bergsteinsson, J.~K. M\o{}ller, P.~Nystrup, {\'O}.~P. P{\'a}lsson,
  D.~Guericke, H.~Madsen, Heat load forecasting using adaptive temporal
  hierarchies, Applied Energy 292 (2021) 116872.

\bibitem{nystrup2021dimensionality}
P.~Nystrup, E.~Lindstr{\"o}m, J.~K. M\o{}ller, H.~Madsen, Dimensionality
  reduction in forecasting with temporal hierarchies, International Journal of
  Forecasting 37~(3) (2021) 1127--1146.

\bibitem{yang2017reconciling}
D.~Yang, H.~Quan, V.~R. Disfani, C.~D. Rodr{\'\i}guez-Gallegos, Reconciling
  solar forecasts: Temporal hierarchy, Solar Energy 158 (2017) 332--346.

\bibitem{edwards1969should}
J.~B. Edwards, G.~H. Orcutt, Should aggregation prior to estimation be the
  rule?, The Review of Economics and Statistics (1969) 409--420.

\bibitem{grunfeld1960aggregation}
Y.~Grunfeld, Z.~Griliches, Is aggregation necessarily bad?, The Review of
  Economics and Statistics (1960) 1--13.

\bibitem{kourentzes2019another}
N.~Kourentzes, D.~Barrow, F.~Petropoulos, Another look at forecast selection
  and combination: Evidence from forecast pooling, International Journal of
  Production Economics 209 (2019) 226--235.

\bibitem{doi:10.1080/01621459.2015.1058265}
J.~Bien, F.~Bunea, L.~Xiao,
  \href{https://doi.org/10.1080/01621459.2015.1058265}{Convex banding of the
  covariance matrix}, Journal of the American Statistical Association 111~(514)
  (2016) 834--845, pMID: 28042189.
\newblock \href
  {http://arxiv.org/abs/https://doi.org/10.1080/01621459.2015.1058265}
  {\path{arXiv:https://doi.org/10.1080/01621459.2015.1058265}}, \href
  {https://doi.org/10.1080/01621459.2015.1058265}
  {\path{doi:10.1080/01621459.2015.1058265}}.
\newline\urlprefix\url{https://doi.org/10.1080/01621459.2015.1058265}

\bibitem{HYNDMAN201616}
R.~J. Hyndman, A.~J. Lee, E.~Wang,
  \href{https://www.sciencedirect.com/science/article/pii/S016794731500290X}{Fast
  computation of reconciled forecasts for hierarchical and grouped time
  series}, Computational Statistics \& Data Analysis 97 (2016) 16--32.
\newblock \href {https://doi.org/https://doi.org/10.1016/j.csda.2015.11.007}
  {\path{doi:https://doi.org/10.1016/j.csda.2015.11.007}}.
\newline\urlprefix\url{https://www.sciencedirect.com/science/article/pii/S016794731500290X}

\bibitem{sagheer2021deep}
A.~Sagheer, H.~Hamdoun, H.~Youness, Deep lstm-based transfer learning approach
  for coherent forecasts in hierarchical time series, Sensors 21~(13) (2021)
  4379.

\bibitem{mancuso2021machine}
P.~Mancuso, V.~Piccialli, A.~M. Sudoso, A machine learning approach for
  forecasting hierarchical time series, Expert Systems with Applications 182
  (2021) 115102.

\bibitem{KOURENTZES2016145}
N.~Kourentzes, F.~Petropoulos,
  \href{https://www.sciencedirect.com/science/article/pii/S0925527315003382}{Forecasting
  with multivariate temporal aggregation: The case of promotional modelling},
  International Journal of Production Economics 181 (2016) 145--153, sI: ISIR
  2014.
\newblock \href {https://doi.org/https://doi.org/10.1016/j.ijpe.2015.09.011}
  {\path{doi:https://doi.org/10.1016/j.ijpe.2015.09.011}}.
\newline\urlprefix\url{https://www.sciencedirect.com/science/article/pii/S0925527315003382}

\bibitem{winkler1992sensitivity}
R.~L. Winkler, R.~T. Clemen, Sensitivity of weights in combining forecasts,
  Operations research 40~(3) (1992) 609--614.

\bibitem{punia2020cross}
S.~Punia, S.~P. Singh, J.~K. Madaan, A cross-temporal hierarchical framework
  and deep learning for supply chain forecasting, Computers \& Industrial
  Engineering 149 (2020) 106796.

\bibitem{miller2020building}
C.~Miller, A.~Kathirgamanathan, B.~Picchetti, P.~Arjunan, J.~Y. Park, Z.~Nagy,
  P.~Raftery, B.~W. Hobson, Z.~Shi, F.~Meggers, the building data genome
  project 2, energy meter data from the ashrae great energy predictor iii
  competition, Scientific data 7~(1) (2020) 1--13.

\bibitem{di2021cross}
T.~Di~Fonzo, D.~Girolimetto, Cross-temporal forecast reconciliation: Optimal
  combination method and heuristic alternatives, International Journal of
  Forecasting (2021).

\bibitem{ledoit2004well}
O.~Ledoit, M.~Wolf, A well-conditioned estimator for large-dimensional
  covariance matrices, Journal of multivariate analysis 88~(2) (2004) 365--411.

\bibitem{schafer2005shrinkage}
J.~Sch{\"a}fer, K.~Strimmer, A shrinkage approach to large-scale covariance
  matrix estimation and implications for functional genomics, Statistical
  applications in genetics and molecular biology 4~(1) (2005).

\bibitem{leprince2020robust}
J.~Leprince, W.~Zeiler, A robust building energy pattern mining method and its
  application to demand forecasting, in: 2020 International Conference on Smart
  Energy Systems and Technologies (SEST), IEEE, 2020, pp. 1--6.

\bibitem{pathak2015constrained}
D.~Pathak, P.~Krahenbuhl, T.~Darrell, Constrained convolutional neural networks
  for weakly supervised segmentation, in: Proceedings of the IEEE international
  conference on computer vision, 2015, pp. 1796--1804.

\bibitem{jia2017constrained}
Z.~Jia, X.~Huang, I.~Eric, C.~Chang, Y.~Xu, Constrained deep weak supervision
  for histopathology image segmentation, IEEE transactions on medical imaging
  36~(11) (2017) 2376--2388.

\bibitem{liu2018constrained}
Y.~Liu, A.~W.~K. Kong, C.~K. Goh, A constrained deep neural network for ordinal
  regression, in: Proceedings of the IEEE conference on computer vision and
  pattern recognition, 2018, pp. 831--839.

\bibitem{drgovna2021physics}
J.~Drgo{\v{n}}a, A.~R. Tuor, V.~Chandan, D.~L. Vrabie, Physics-constrained deep
  learning of multi-zone building thermal dynamics, Energy and Buildings 243
  (2021) 110992.

\bibitem{marquez2017imposing}
P.~M{\'a}rquez-Neila, M.~Salzmann, P.~Fua, Imposing hard constraints on deep
  networks: Promises and limitations, arXiv preprint arXiv:1706.02025 (2017).

\bibitem{CHEN201369}
D.~Chen, H.~W. Chen,
  \href{https://www.sciencedirect.com/science/article/pii/S2211464513000328}{Using
  the k{\"o}ppen classification to quantify climate variation and change: An
  example for 1901-2010}, Environmental Development 6 (2013) 69--79.
\newblock \href {https://doi.org/https://doi.org/10.1016/j.envdev.2013.03.007}
  {\path{doi:https://doi.org/10.1016/j.envdev.2013.03.007}}.
\newline\urlprefix\url{https://www.sciencedirect.com/science/article/pii/S2211464513000328}

\bibitem{knmi}
KNMI, \href{https://www.daggegevens.knmi.nl/klimatologie/uurgegevens}{Hourly
  values of weather stations} (Aug 2021).
\newline\urlprefix\url{https://www.daggegevens.knmi.nl/klimatologie/uurgegevens}

\bibitem{mullner2011modern}
D.~M{\"u}llner, Modern hierarchical, agglomerative clustering algorithms, arXiv
  preprint arXiv:1109.2378 (2011).

\bibitem{reshef2011detecting}
D.~N. Reshef, Y.~A. Reshef, H.~K. Finucane, S.~R. Grossman, G.~McVean, P.~J.
  Turnbaugh, E.~S. Lander, M.~Mitzenmacher, P.~C. Sabeti, Detecting novel
  associations in large data sets, science 334~(6062) (2011) 1518--1524.

\bibitem{miller2019s}
C.~Miller, What's in the box?! towards explainable machine learning applied to
  non-residential building smart meter classification, Energy and Buildings 199
  (2019) 523--536.

\bibitem{scikit-learn}
F.~Pedregosa, G.~Varoquaux, A.~Gramfort, V.~Michel, B.~Thirion, O.~Grisel,
  M.~Blondel, P.~Prettenhofer, R.~Weiss, V.~Dubourg, J.~Vanderplas, A.~Passos,
  D.~Cournapeau, M.~Brucher, M.~Perrot, E.~Duchesnay, Scikit-learn: Machine
  learning in {P}ython, Journal of Machine Learning Research 12 (2011)
  2825--2830.

\bibitem{hewamalage2022forecast}
H.~Hewamalage, K.~Ackermann, C.~Bergmeir, Forecast evaluation for data
  scientists: Common pitfalls and best practices, arXiv preprint
  arXiv:2203.10716 (2022).

\bibitem{chollet2015keras}
F.~Chollet, et~al., \href{https://github.com/fchollet/keras}{Keras} (2015).
\newline\urlprefix\url{https://github.com/fchollet/keras}

\end{thebibliography}

\end{document}